\documentclass[iicol,sn-mathphys]{sn-jnl}

\usepackage{graphicx}%
\usepackage{multirow}%
\usepackage{amsmath,amssymb,amsfonts}%
\usepackage{amsthm}%
\usepackage{mathrsfs}%
\usepackage[title]{appendix}%
\usepackage{xcolor}%
\usepackage{textcomp}%
\usepackage{manyfoot}%
\usepackage{booktabs}%
\usepackage{algorithm}%
\usepackage{listings}%
\usepackage{tabularx}
\usepackage{hyperref}       
\usepackage{algorithmic}
\usepackage{array}
\usepackage[caption=false,font=normalsize,labelfont=sf,textfont=sf]{subfig}
\usepackage{textcomp}
\usepackage{stfloats}
\usepackage{url}
\usepackage{verbatim}
\usepackage{graphicx}
\usepackage{cite}
\usepackage{booktabs}       
\usepackage{nicefrac}       
\usepackage{microtype}      
\usepackage{color}
\usepackage{colortbl}
\usepackage{multicol}
\usepackage{diagbox}
\usepackage{wrapfig}
\usepackage{multirow}
\usepackage{makecell}
\usepackage{subcaption}
\usepackage{wrapfig}
\usepackage{cleveref}
\usepackage{caption}
\usepackage{newfloat}
\definecolor{tableColor}{HTML}{e9f1f6}
\newcolumntype{a}{>{\columncolor{tableColor}}c}

\theoremstyle{thmstyleone}%
\theoremstyle{thmstyletwo}%

\theoremstyle{thmstylethree}%

\begin{document}

\title[Article Title]{ResAD++: Towards Class Agnostic Anomaly Detection via Residual Feature Learning}


\author[1]{\fnm{Xincheng} \sur{ Yao}}\email{i-dover@sjtu.edu.cn}

\author[1]{\fnm{Chao} \sur{Shi}}\email{shichaostone@sjtu.edu.cn}

\author[2]{\fnm{Muming} \sur{Zhao}}\email{mumingzhao@bjfu.edu.cn}

\author[1]{\fnm{Guangtao} \sur{Zhai}}\email{zhaiguangtao@sjtu.edu.cn}

\author*[1]{\fnm{Chongyang} \sur{Zhang}}\email{sunny\_zhang@sjtu.edu.cn}

\affil*[1]{\orgname{School of Information Science and Electronic Engineering, Shanghai Jiao Tong University}, \orgaddress{\state{Shanghai}, \country{China}}}
\affil[2]{\orgname{School of Information Science and Technology, Beijing Forestry University}, \orgaddress{\state{Beijing}, \country{China}}}




\abstract{This paper explores the problem of class-agnostic anomaly detection (AD), where the objective is to train one class-agnostic AD model that can generalize to detect anomalies in diverse new classes from different domains without any retraining or fine-tuning on the target data. When applied for new classes, the performance of current single- and multi-class AD methods is still unsatisfactory. One fundamental reason is that representation learning in existing methods is still class-related, namely, feature correlation. Feature correlation represents that the features of each class have many class-related attributes to the class, resulting in the model learned on one class relying on class-related representations, thereby being hard to adapt to other classes. To address this issue, we propose residual features and construct a simple but effective framework, termed ResAD. Our core insight is to learn the residual feature distribution rather than the initial feature distribution. Residual features are formed by matching and then subtracting normal reference features. In this way, we can effectively realize feature decorrelation. Even in new classes, the distribution of normal residual features would not remarkably shift from the learned distribution. In addition, we think that residual features still have one issue: scale correlation. Scale correlation refers that features of different classes may still have significant differences in scale, namely, the numerical value scales in the features of different classes may be remarkably different. To this end, we propose a feature hypersphere constraining approach, which learns to constrain initial normal residual features into a spatial hypersphere for enabling the feature scales of different classes as consistent as possible. Furthermore, we propose a novel log-barrier bidirectional contraction OCC loss and vector quantization based feature distribution matching module to enhance ResAD, leading to the improved version of ResAD (ResAD++). Comprehensive experiments on eight real-world AD datasets demonstrate that our ResAD++ can achieve remarkable AD results when directly used in new classes, outperforming state-of-the-art competing methods and also surpassing ResAD. The code is available at \url{https://github.com/xcyao00/ResAD}.}




\maketitle

\section{Introduction}
\label{sec:introduction}

\begin{figure}[ht]
    \centering
    \includegraphics[width=1.0\linewidth]{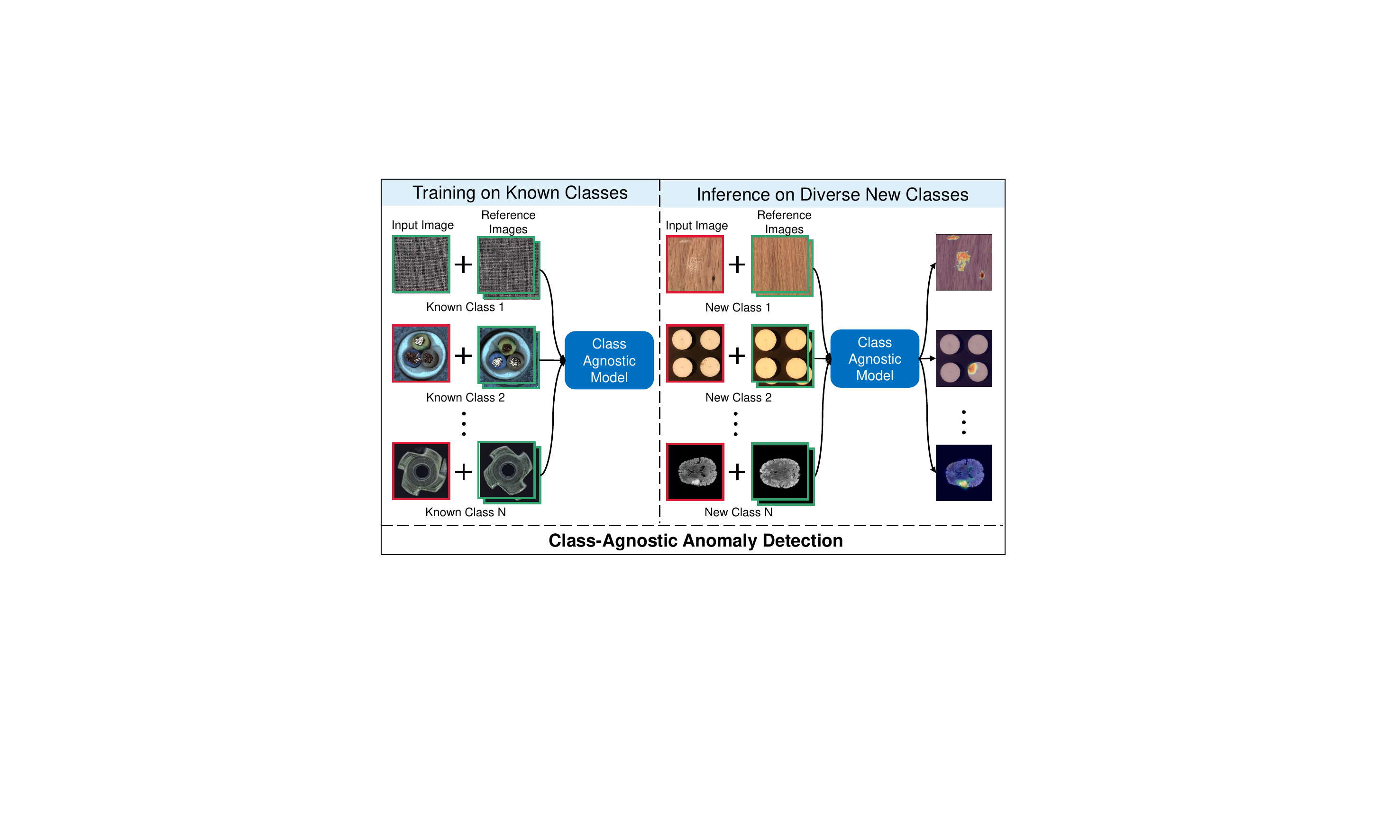}
    \caption{\textbf{Intuitive illustration of class-agnostic anomaly detection}. We aim to train a class-agnostic AD model that can be directly applied to detect anomalies in new classes with only few-shot normal samples as reference. Class means the category of the object in the image. For industrial scenarios, it refers to the industrial product category, \emph{e.g.}, carpet, cable, etc. For medical analysis, it refers to the body organ category, \emph{e.g.}, head, retina, etc.}
    \label{fig:motivation1}
\end{figure}

Anomaly detection (AD) is a widely studied machine learning task that aims to distinguish an instance that substantially deviates from the normal samples in the dataset. In computer vision, image-based anomaly detection typically also requires further localizing those anomalous image regions. In real-world applications, anomaly detection has received widespread attention in diverse domains, such as industrial defect inspection \cite{MVTec, MVTec3D, VisA}, video surveillance \cite{UBnormal, RealWorldAD}, medical lesion detection \cite{Survey1, Survey2}, and road anomaly detection \cite{RoadAD1, RoadAD2}. Due to the scarcity of anomalies and diversity of normal classes (see Fig.\ref{fig:motivation1} for the meaning of class), current anomaly detection studies are mainly devoted to unsupervised one-for-one learning, \emph{i.e.}, learning one specific AD model with only anomaly-free samples for each class. Most of the previously popular AD methods follow this paradigm, such as reconstruction-based methods \cite{SSIM, MemoryAE, DFR, AnoGAN, GANomaly, Intra, SSPCAB, HETMM}, one-class-classification based methods \cite{deepSVDD, SAD, FCDD, PatchSVDD, MS-PatchSVDD}, embedding-based methods \cite{PaDiM, PatchCore, DifferNet, CFLOW, FastFLOW}, and knowledge distillation approaches \cite{STAD, MKD, STPM, RDAD, RDAD+, AST}.

Despite these methods having achieved remarkable detection performance on various AD benchmarks, applying anomaly detection algorithms in real-world scenarios still confronts many challenges. A critical challenge is that there are usually diverse classes, and new classes are continually emerging. The previous single-class AD methods and also multi-class AD methods \cite{UniAD, OmniAL, PMAD, HGAD, DiAD, MambaAD} are still insufficient to satisfy the requirements of real-world applications. Multi-class AD aims to learn one AD model for multiple classes simultaneously, but the learned model can't be directly applied to new classes and still requires retraining. In some privacy-sensitive scenarios, a more serious problem is that these methods become infeasible as retraining on the target data is not allowed due to data privacy issues. Therefore, class-agnostic ability is a critical issue in the AD community, but it still hasn't been well studied in current AD literatures. To tackle the critical AD challenge, this paper aims to study an academy-valuable and application-required task: \textbf{class-agnostic anomaly detection}. As shown in Fig.\ref{fig:motivation1}, \emph{one class-agnostic model is trained with samples from multiple known classes, and the objective is to generalize to detect anomalies in new\footnote{In this paper, we call the classes in training as known classes, others are called as new (or novel, unknown) classes.} classes. When used for new class detection, only few-shot new class normal samples are required, without retraining or any fine-tuning on the target data}. Compared to multi-class AD, class-agnostic AD is more challenging because this task also requires one AD model for multiple classes, and even these classes are not trained.

 Nonetheless, solving such a task is quite challenging. The performance of current single-class and also multi-class AD models usually drops dramatically when used for new classes (see Tab.\ref{tab:main_results}). We attribute this phenomenon to one issue: \textbf{class-related feature representation, \emph{a.k.a,} feature correlation}. Feature correlation represents that the features of each class generated by well-trained neural networks usually have many class-related attributes \cite{DeepRepresentation}. As class-related attributes are an abstract concept (hard to intuitively visualize), we explain them through specific images. For example, in the left part of Fig.\ref{fig:motivation1}, the weaving texture is one typical attribute for the known class 1. For the known class 2, three different colored insulators, silver and brown wires, and blue cable cross-sections are typical attributes. For the known class 3, four counterclockwise rotating gears and bronze with black colored surface are typical attributes. After feature extraction, the class-related attributes are embedded into features, which are more abstract and hard to describe in language. We can only plot the feature distribution for visualization. As different classes have distinctive class-related attributes, the feature distributions of different classes will be significantly different (please see Fig.\ref{fig:vis_feature_decorrelation}(a)). In a nutshell, class-related attributes are typical to the class and distinctive from other classes, representing the discriminative characteristics of the class. This can also explain the failure of previous methods when dealing with novel classes. Because previous methods didn’t consider eliminating the significant ``normal-to-normal'' discrepancy. The ``normal-to-normal'' discrepancy causes normal in new classes will also be misjudged as abnormal (see Fig.\ref{fig:vis_results}). 
 
\begin{figure}[ht]
    \centering
    \includegraphics[width=1.0\linewidth]{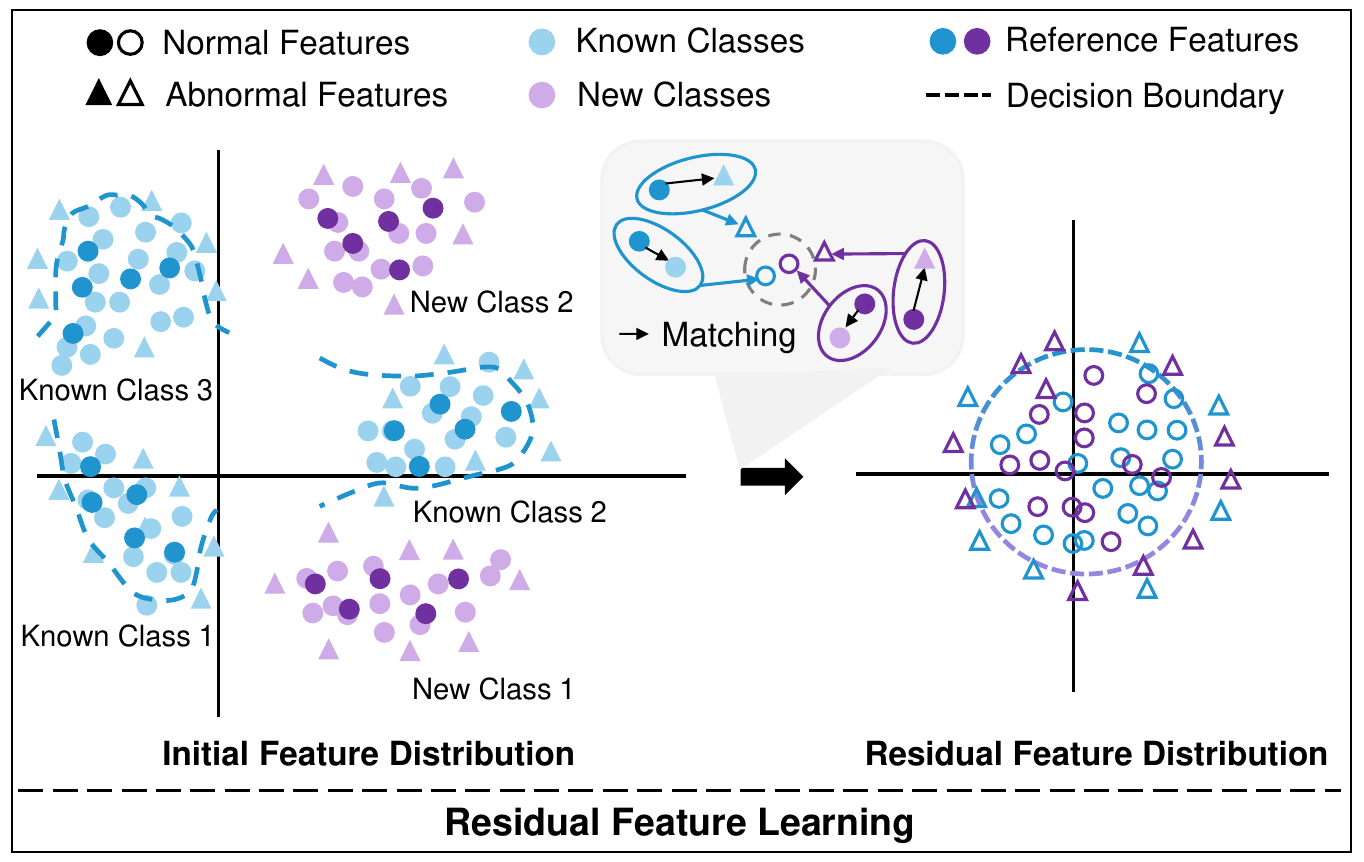}
    \caption{\textbf{Conceptual illustration of residual features}. The residual feature space has fewer variations compared to the initial feature space. The decision boundary of the residual feature distribution can more effectively distinguish anomalies in new classes, rather than treating most features of new classes as anomalies.}
    \label{fig:motivation2}
\end{figure}

 To address the feature correlation issue, our core insight is to utilize residual features to effectively realize feature decorrelation. The residual operation includes two steps: matching and subtracting. We will first match the nearest normal reference feature to each input feature. As class-related attributes can also be embedded in normal reference features\footnote{For example, in Fig.\ref{fig:motivation1}, each normal reference sample of the known class 2 will also contain three insulators (class-related) with different colors, otherwise it's not normal.}, the closer the two features are, usually the more similar the embedded attributes will be. Then, the matching process at the patch level can be considered as matching the most similar class-related attributes to each input feature. Therefore, by further subtracting, it is highly probabilistic that the class-related components in the initial features will be mutually eliminated. From the perspective of feature distribution, it can be imagined that residual features will be distributed in a relatively fixed origin-centered region (see Fig.\ref{fig:motivation2} and Fig.\ref{fig:vis_feature_decorrelation}(b)). As shown in Fig.\ref{fig:motivation2}, the main merit of normal residual features is: even in new classes, the distribution of normal residual features would not remarkably shift from the learned distribution. Regardless of classes, larger residual values are expected for abnormal features than normal features. Therefore, we think that residual features can be regarded as class-invariant\footnote{Strictly speaking, the residual features are not fully invariant, while the variations are significantly smaller.} representations compared to the significantly variant initial features.

 Based on residual features, we propose a simple but effective class-agnostic AD framework, termed ResAD (\emph{i.e.}, \textbf{Res}idual Feature Learning based Class-Agnostic \textbf{A}nomaly \textbf{D}etection). ResAD is based on one key insight: residual feature learning, and consists of two key designs: feature hypersphere constraining and feature distribution estimating. First, we employ a pre-trained feature extractor to generate normal reference features from few-shot normal reference samples. Each input feature will match the nearest normal reference feature and subtract it to form the residual feature.
 
 Second, we think that residual features still have one issue: \textbf{class-related scale}\footnote{Note that we use the ``scale'' to represent the size of numerical values in features or the size of feature norms, rather than object scale.} \textbf{variation, \emph{a.k.a}, scale correlation}. Scale correlation refers that features of different classes may also have significant differences in scale, namely, the numerical value scales in the features of different classes may be remarkably different. This can lead to difficulty in obtaining a unified normal-abnormal decision boundary for different classes, \emph{i.e.}, the scales of decision boundaries in different classes may be significantly different, a good decision boundary in one class may be poor in another class (please see detailed explanations in Sec.\ref{sec:feature_constraintor}). To address the scale correlation issue, we take the idea from one-class-classification learning \cite{SVDD, FCDD} and propose a feature hypersphere constraining approach to realize scale decorrelation. In the scale decorrelation module, the network will learn to constrain initial normal residual features into a spatial hypersphere for enabling the feature scales of different classes as consistent as possible. 
 
 Third, with the hypersphere-constrained residual features, we can easily utilize a feature distribution estimator \cite{realNVP} to learn and estimate the normal residual feature distribution, anomalies can be recognized as out-of-distribution. Furthermore, to further reduce distribution mismatch between testing feature distribution and training feature distribution, we propose a vector quantization (VQ) based feature distribution matching approach to inject training distribution into the test samples. In this way, the residual features of new classes are more likely to match certain learned representations, thus being more accurately discriminated. For new classes, as the residual features have fewer variations or are covered by the learned distribution, the whole framework is more class-agnostic. 

 The idea of residual features was initially introduced in our recent NeurIPS 2024 spotlight paper \cite{ResAD}. In this work, we mainly make four improvements to ResAD: (1) We rethink the causes of why existing methods are poorly class-agnostic and propose a more essential explanation: feature correlation. The deeper explanation can better inspire the solution: decorrelation. More importantly, we further numerically evaluate the effectiveness of our method for achieving feature decorrelation and scale decorrelation by statistical values (see Tab.\ref{tab:feature_decorrelation_statistics} and Tab.\ref{tab:scale_decorrelation_statistics}). (2) In the feature hypersphere constraining module, we propose a novel log-barrier bidirectional contraction OCC loss, which is more robust and can achieve better results. (3) In the feature distribution estimating module, we modify the maximum likelihood loss and employ a new loss for optimization. This improvement provides us with a new anomaly scoring way, which is better for AD results. (4) We propose a novel vector quantization based feature distribution matching module, which can make the testing distribution more consistent with the training distribution. Aside from these method changes, we conduct more comprehensive experiments, including comparison on more datasets, adding a new 8-shot setting, comparison with full-shot trained models, and more extensive ablation studies, to clearly illustrate the advantage of our method. The improved method is called as ResAD++ for distinguishing from the name in the conference paper. Overall, the performance of ResAD++ is better than ResAD in \cite{ResAD}.
 
 In summary, we make the following main contributions:

1. We study the class-agnostic anomaly detection task to evaluate the class-agnostic ability of AD methods in identifying anomalies from novel classes without retraining or fine-tuning. To the best of our knowledge, relevant research works about class-agnostic AD are still relatively lacking in the AD community.

2. Based on our previous work, ResAD \cite{ResAD}, we further extend the original version of ResAD to ResAD++ with novel improvements, including log-barrier bidirectional contraction OCC loss, VQ-based feature distribution matching module, etc. We think that the idea of residual feature learning is not specific to the proposed framework and can provide inspiration for subsequent class-agnostic AD works.

3. Comprehensive experiments on eight AD datasets are performed to evaluate our ResAD++'s class-generalization ability. The results show that our ResAD++ can achieve remarkable AD results when directly used in new classes, outperforming state-of-the-art methods and also surpassing ResAD. With only 8-shot normal samples as reference, ResAD++ can achieve 98.6\% image-level AUROC and 96.7\% pixel-level AUROC on MVTecAD. 

\section{Related Work}
\label{sec:related_work}

\textbf{Single-class Anomaly Detection.} Most AD methods follow the one-for-one paradigm. \emph{Reconstruction-based methods} are the most popular AD methods. These methods hold the insight that models trained by normal samples would fail in abnormal image regions. Many previous works utilize auto-encoders \cite{SSIM, MemoryAE, DFR}, variational auto-encoders \cite{VAE1} and generative adversarial networks \cite{AnoGAN, GANomaly} to encode and reconstruct normal data. Other methods \cite{Intra, SSPCAB, PMAD} accomplish anomaly detection by inpainting, where image patches are masked randomly. Then, neural networks are trained to predict the masked patches. Distillation-based AD methods \cite{STAD} can also be considered as belonging to the reconstruction type. These methods train student networks to reconstruct the representations of teacher networks on normal samples, and the assumption is that the student would fail in abnormal features. Recent works mainly focus on feature pyramid \cite{MKD, STPM}, reverse distillation \cite{RDAD, RDAD+}, and asymmetric distillation \cite{AST}. 

\emph{OCC-based methods} belong to another type of classical AD modeling methods. The earliest works are mainly to extend the OCC (\emph{i.e.}, one-class-classification) models such as OC-SVM \cite{OneclassSVM} or SVDD \cite{SVDD, deepSVDD} for anomaly detection. Recently, in \cite{PatchSVDD}, a patch-based SVDD that contains multiple cores rather than a single core in DeepSVDD \cite{deepSVDD} is proposed to enable anomaly localization. DeepSAD \cite{deepSVDD} is the first semi-supervised OCC-based AD framework that extends DeepSVDD to utilize a few abnormal samples during training. FCDD proposed in \cite{FCDD} further extends DeepSAD based on the pseudo-Huber loss in \cite{HSC} to support anomaly localization. In FCDD \cite{FCDD}, anomaly maps are generated by the proposed Fully Convolutional Data Description combined with receptive field upsampling. In \cite{MS-PatchSVDD}, the authors further extend the PatchSVDD \cite{PatchSVDD} model by the proposed multi-scale patch-based representation learning method.

\emph{Embedding-based methods} recently show state-of-the-art performance. These methods mainly rely on good feature representations and assume that abnormal features are usually far from the normal clusters. Most superior methods \cite{SPADE, DeepKNN, PaDiM, PaDiM1, PatchCore} utilize ImageNet pre-trained networks for feature extraction. PaDiM \cite{PaDiM} extracts pre-trained features to model Multivariate Gaussian distribution and then utilizes Mahalanobis distance to measure the anomaly scores. PatchCore \cite{PatchCore} extends on this line by utilizing locally aggregated features and introducing a maximally representative memory bank of normal features. However, industrial images generally have an obvious distribution shift from ImageNet. To better account for the distribution shift, subsequent adaptations should be done. The normalizing flow based methods \cite{DifferNet, CFLOW, CSFLOW, FastFLOW} are proposed to transform the pre-trained feature distribution into a latent Gaussian distribution, and thus can better learn the normal data distribution.

\textbf{Multi-class Anomaly Detection.} Recently, some researchers have attempted to jump out from the one-for-one paradigm and studied how to design one multi-class AD model to accomplish anomaly detection for multiple classes simultaneously. UniAD \cite{UniAD}, OmniAL \cite{OmniAL}, and HVQ-Trans \cite{HVQ-Trans} are early proposed methods for this new direction. UniAD is a transformer-based reconstruction model and mainly based on three improvements, \emph{layer-wise query decoder}, \emph{neighbor masked attention}, and \emph{feature jittering strategy}, to address the “identical shortcut” issue to achieve multi-class AD. OmniAL is a unified CNN framework with anomaly synthesis, reconstruction, and localization improvements. To prevent identical reconstruction, OmniAL trains the model with panel-guided synthetic anomaly data rather than directly using normal data. HVQ-Trans follows UniAD, which mitigates the shortcut learning issue via a vector quantization mechanism. Unlike the above three reconstruction-based methods, HGAD \cite{HGAD} is a normalizing flow (NF) based AD model. The authors find that popular NF-based AD methods may fall into a “homogeneous mapping” issue when used for multi-class AD, and correspondingly propose a novel hierarchical Gaussian mixture normalizing flow modeling method to address this issue. 

DiAD \cite{DiAD}, RLR \cite{RLR}, MambaAD \cite{MambaAD}, Dinomaly \cite{Dinomaly}, and INP-Former \cite{INP-Former} are more recent works. DiAD investigates a multi-class AD framework based on diffusion models, introducing a semantic-guided network to ensure the consistency of reconstructed image semantics. RLR utilizes learnable reference representations to compel the model to learn normal feature patterns explicitly, thereby preventing the model from succumbing to the shortcut issue. MambaAD is the first exploration work to employ state space models (SSM) to address multi-class AD, introducing a LSS module with hybrid state space (HSS) blocks and multi-kernel convolutions to effectively capture both long-range and local information. In Dinomaly, the authors follow the ``less is more'' philosophy to propose a pure Transformer-based AD framework without relying on complex designs or specialized tricks. To avoid the student overly mimicking the teacher, Dinomaly employs three simple components: \emph{noisy bottleneck}, \emph{linear attention}, and \emph{loose reconstruction} to weaken the overfitting of the network during training. INP-Former proposes the intrinsic normal prototypes (INPs) concept, which are adaptively extracted from the test image itself. These INPs can provide more concise and well-aligned prototypes to the anomalies than those learned from training data. With the guidance of INPs, the decoder can more accurately reconstruct normal regions and effectively suppress the reconstruction of anomalous regions. Furthermore, the improved method, INP-Former++ \cite{INP-Former++}, utilizes our residual feature learning to amplify the discriminative boundary between normal and abnormal features for further performance improvement.

\textbf{Unified Anomaly Detection with Few-shot Learning.} We think that few-shot AD methods are also valuable efforts to achieve class-agnostic anomaly detection. Few-shot AD methods can be regarded as indirectly achieving class-agnostic by only utilizing few-shot normal samples to construct AD models. Distance-based approaches such as SPADE \cite{SPADE}, PaDiM \cite{PaDiM} and PatchCore \cite{PatchCore} can be adapted to address few-shot AD by only making use of few-shot samples to calculate distance-based anomaly scores without training. RegAD \cite{RegAD} proposes to train a feature registration network to align input images and follows \cite{PaDiM} to model Multivariate Gaussian distribution with few-shot normal samples. Recently, the CLIP-based AD methods, including WinCLIP \cite{WinCLIP} and AprilGAN \cite{VAND}, show better few-shot and even zero-shot AD performance. WinCLIP proposes multi-scale aggregation to construct small-scale and mid-scale feature memories. AprilGAN introduces extra linear layers to map image features to the text feature space. They both employ a text prompt ensemble strategy to obtain the language-guided anomaly map. AnoVL \cite{AnoVL} further introduces to replace the QKV attention with V-V attention for enhancing the local visual semantics. Subsequently, many works actively attempt to accomplish zero-shot anomaly detection based on the CLIP model \cite{CLIP}, such as AnomalyCLIP \cite{AnomalyCLIP}, CLIP-AD \cite{CLIP-AD}, ClipSAM \cite{ClipSAM}, and FiLo \cite{FiLo}. Although different from the general definition of zero-shot AD, in MuSc \cite{MuSc}, the authors also call their method as zero-shot AD, which detects anomalies by directly excavating normal information from unlabeled test images, without the need for normal training images.

   Different from class-agnostic AD, few-shot AD mainly focuses on how to effectively utilize few-shot normal samples to construct AD models. Some dedicated modules may be introduced to handle the few-shot normal samples. These methods still need to remodel or retrain in new classes based on few-shot normal samples, \emph{e.g.}, PatchCore needs to reconstruct the coreset, and RegAD needs to remodel the Multivariate Gaussian distribution for new classes. Therefore, few-shot AD methods are still class-reliant essentially. The CLIP-based methods can be seen as class-agnostic, as these methods can obtain anomaly maps by aligning vision features with text features without remodeling in new classes. However, they rely on the visual-language comprehension ability of CLIP and handcrafted text prompts about anomalies (new classes may require new specific text prompts), making them difficult to generalize to anomalies in diverse classes. They may fail to work well when the text prompts cannot capture the desired anomaly semantics, especially for these CLIP-based zero-shot AD methods. Because these methods heavily rely on language descriptions of anomalies and don't effectively utilize normal samples. However, normal information is crucial for detecting anomalies, and few-shot normal samples are easy to acquire. 

 More recently, the idea of in-context residuals in InCTRL \cite{InCTRL} is very similar to ours. But our method has obvious differences with InCTRL in the definition and utilization of residuals. (1) The definition of residuals in InCTRL is based on feature distances. We think that residual distances in InCTRL can limit the range of residual representation (as the cosine similarity is in [-1,1]). In contrast, our residual features don’t limit the range of residual representation and can retain the feature properties. In high-dimensional feature space, we can establish better decision boundaries between normal and abnormal. (2) InCTRL is to train a binary classification network based on residual distance maps, which can only output an image-level anomaly score. Our method is to learn the distribution of residual features, an anomaly score can be estimated for each feature, thus can be used to locate anomalies. (3) Our residual features can be easily applied to other AD methods by simply replacing the original features, while the residuals in InCTRL are more dependent on the whole method and not easily applicable to other AD methods. (4) Our ResAD++ can remarkably surpass InCTRL on multiple datasets (see Tab.\ref{tab:main_results}).

\section{Method}

\begin{figure*}[ht]
    \centering
    \includegraphics[width=1.0\linewidth]{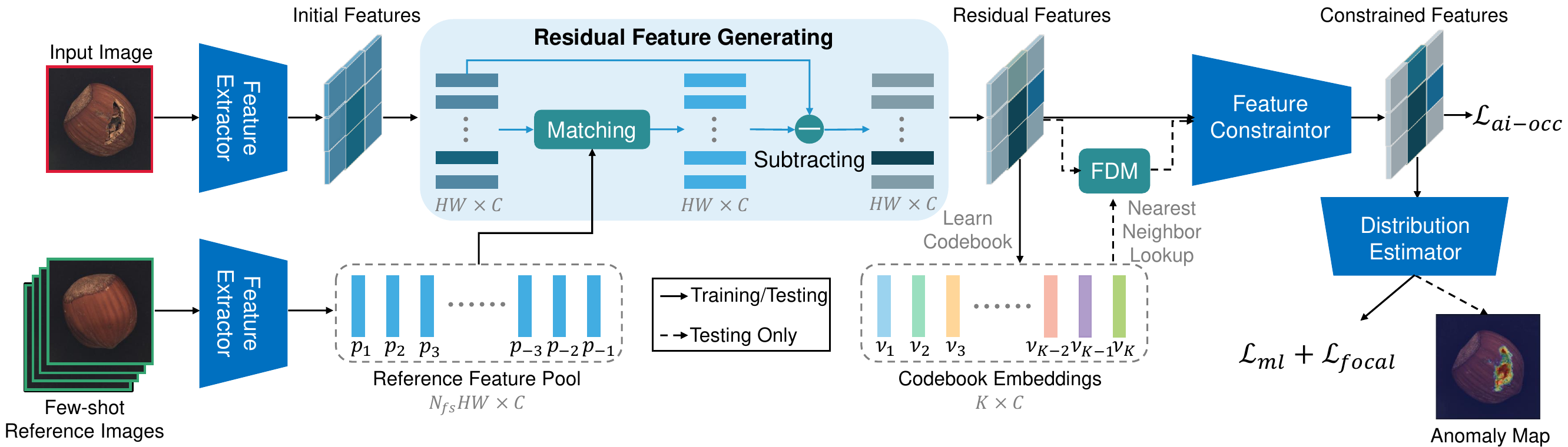}
    \caption{\textbf{Framework overview}. First, few-shot normal reference samples are fed into a pre-trained \emph{Feature Extractor} to obtain normal reference features. These features are stored in reference feature pools, we only show one pool. For each input image, we utilize the same \emph{Feature Extractor} to extract multi-layer feature maps, the figure only shows one layer. Each initial feature will match the nearest normal reference feature and subtract it to form the residual feature. The codebook is learned during training by vector quantization (VQ) technique to effectively represent the training residual feature distribution. The \emph{Feature Distribution Matching (FDM)} module is only utilized during testing for further making the testing data distribution more consistent with the training data distribution (see Sec.\ref{sec:fdm}). Then, a \emph{Feature Constraintor} is utilized to transform the normal residual features into a constrained spatial hypersphere. Finally, we employ a normalizing flow model as the \emph{Feature Distribution Estimator} to learn and estimate the residual feature distribution. During testing, the \emph{Feature Distribution Estimator} can be used to estimate log-likelihoods for input features, anomalies usually have smaller log-likelihoods, thereby being located. Note that training samples and testing samples belong to different classes.}
    \label{fig:framework}
\end{figure*}

\textbf{Problem Statement.} The objective of class-agnostic anomaly detection is to train one class-agnostic AD model that can still work well for detecting anomalies on novel classes from diverse application domains without any retraining or fine-tuning on the target data. Thus, the classes in the training set are assumed to be different from the classes in the test sets. Formally, let $\mathcal{I}_{train} = \mathcal{I}^n \cup \mathcal{I}^a$ be an auxiliary training dataset with normal images and some anomalies (\emph{i.e.}, anomalies that exist in the training set should also be effectively utilized), where $\mathcal{I}^n = \{I^n_i\}_{i=1}^{N_0}$ and $\mathcal{I}^a = \{I^a_j\}_{j=1}^{M_0}$ indicate the collection of normal samples and abnormal samples. As for testing, the model evaluation is conducted on a collection of other AD datasets ($\mathcal{T} = \{\mathcal{I}^{test}_1, \mathcal{I}^{test}_2, \dots, \mathcal{I}^{test}_T\}$) except the training dataset. The classes in the test set are drawn from unknown classes $\mathcal{C}_u$ that are different from the known classes $\mathcal{C}_k$ in the training set. Then the goal is to learn a class-agnostic model $\mathcal{M} : \mathcal{I} \rightarrow \mathbb{R}$ that is trained with known classes $\mathcal{C}_k$ and can directly adapt to unknown classes $\mathcal{C}_u$ with only few-shot (\emph{e.g.}, 4) normal samples as reference. Please note that the reference samples used during testing are not available in any way during the training of the class-agnostic AD model.

\textbf{Overview.} Towards class-agnostic anomaly detection, we innovatively propose the residual feature learning approach and construct a simple framework, ResAD++. The framework is illustrated in Fig.\ref{fig:framework}. The detailed workflow description of ResAD++ is in the caption of Fig.\ref{fig:framework}. Our proposed residual feature learning approach consists of four parts: \emph{residual feature generating}, \emph{feature hypersphere constraining}, \emph{feature distribution estimating}, and \emph{VQ-based feature distribution matching}. These modules will be described below in sequence.

\subsection{Residual Feature Generating}
\label{sec:residual_features}
Residual feature learning is our core insight for solving class-agnostic anomaly detection. In this subsection, we describe how to generate residual features. For any input image $I_i \in \mathbb{R}^{H_0\times W_0\times 3}$, we follow the common practice of previous AD methods to employ a pre-trained feature extraction network $\phi$ to extract features from different layers. Formally, we define $L$ as the total number of layers for use. The feature map from layer $l \in \{1,2,\dots,L\}$ is denoted as $\phi^{l}(I_i) \in \mathbb{R}^{H_l\times W_l \times C_l}$, where $H_l$, $W_l$, and $C_l$ are the height, width, and channel dimension of the feature map. For a feature vector $x_{h,w}^l = \phi^{l}(I_i)_{h,w} \in \mathbb{R}^{C_l}$ at layer $l$ and location $(h,w)$, we will match it with the nearest normal reference feature from the corresponding reference feature pool, and then convert it into the residual feature. The details are described in the following:

\textbf{Reference Feature Pools.} The reference feature pools are utilized to store some normal features as reference. For new classes, we will provide few-shot normal samples (\emph{i.e.}, randomly selected and then fixed) as reference. The pre-trained network $\phi$ will also extract multi-layer features for these normal reference images, then the extracted features are sent into the feature pools as reference features. For $l$th layer, the $l$th reference feature pool is composed of $\mathcal{P}_l = \{x_{h,w}^{l,i}|h\in \{1,\dots,H_l\}, w\in \{1,\dots,W_l\}, l\in \{1,\dots,L\}, i\in \{1,\dots,N_{fs}\}\}$, where $i$ denotes the $i$th normal reference sample, the $N_{fs}$ is the number of normal reference samples.

\textbf{Residual Features.} For each initial feature $x_{h,w}^l$, we can search the nearest nominal reference feature $x_n^* = \mathop{{\rm argmin}}_{x\in\mathcal{P}_l}||x-x_{h,w}^l||_2$ from the $l$th reference feature pool $\mathcal{P}_l$. Then, we define the residual representation of $x_{h,w}^l$ to its closest normal reference feature as:
\begin{equation}
    x_{h,w}^{l,r} = x_{h,w}^l - x_n^*
\end{equation}

\begin{figure}[ht]
    \centering
    \includegraphics[width=1.0\linewidth]{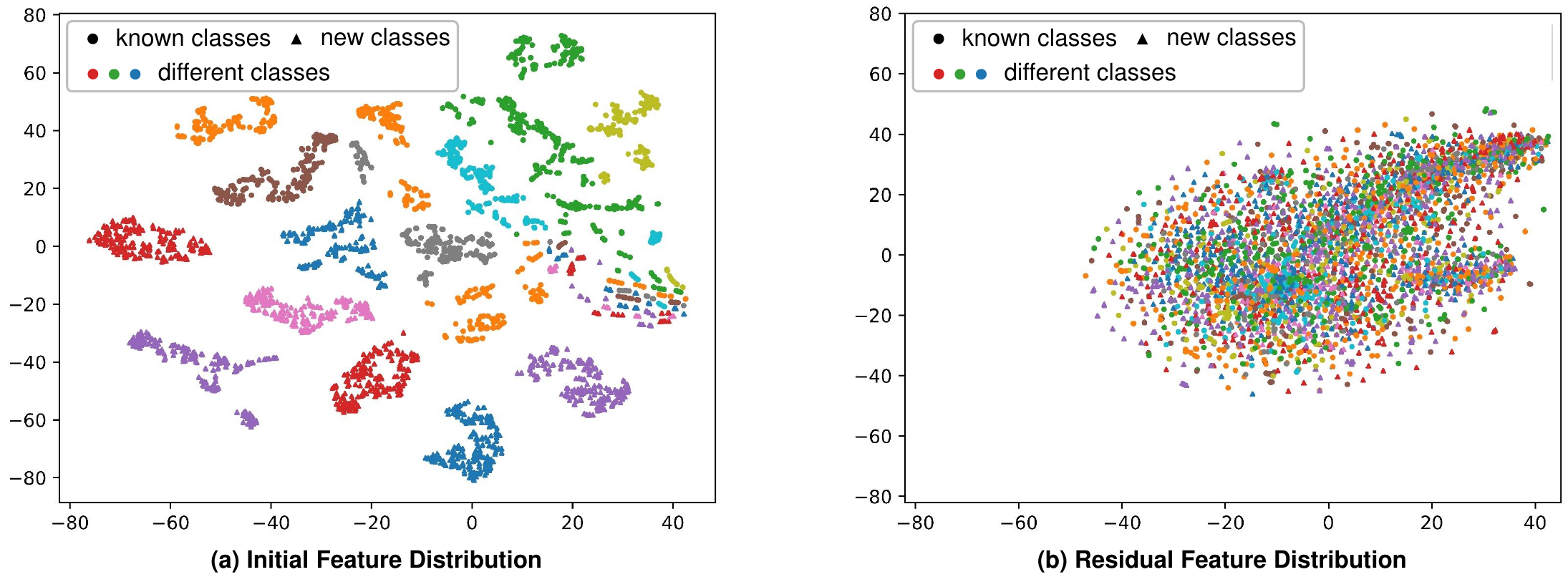}
    \caption{\textbf{Feature t-SNE visualization on the MVTecAD dataset}. (a) In the initial feature space, the features from different classes are significantly different. (b) In the residual feature space, even the residual feature distribution of unknown classes would not remarkably shift from the known distribution. Note that in (a) and (b), we only show normal features and use different colors to represent different classes.}
    \label{fig:vis_feature_decorrelation}
\end{figure}

\textbf{Why are residual features more class-agnostic compared to initial features}? We think that residual features can effectively realize feature decorrelation (they are generated by matching and then subtracting). From the principles of representation learning, we know that features of each class generated by well-trained neural networks usually have many class-related attributes to the class for distinguishing from other classes \cite{DeepRepresentation}. The ``class-related'' means these attributes are typical to the class and distinctive from other classes, representing the discriminative characteristics of the class (please see more explanations in the introduction). Thus, features from different classes are usually located in different feature domains \cite{Domain}. As class-related attributes can also be embedded in normal reference features, the closer two features are, usually the more similar the embedded attributes will be. Then, the matching process at the patch level can be seen as matching the most similar class-related attributes to each query feature. Therefore, by subtracting, it is highly probabilistic that the class-related components in the initial features will be mutually eliminated. It can be imagined that residual features will be distributed in a relatively fixed region centered around the origin. A prominent merit of residual features for class-agnostic anomaly detection is that they have fewer variations in unknown classes. Therefore, even in new classes, the distribution of normal residual features would not remarkably shift from the learned distribution. In Fig.\ref{fig:vis_feature_decorrelation}, the t-SNE visualization of initial feature and residual feature distributions also empirically validates our above explanations. In addition, residual features are also beneficial to highlight the discrepancy between normals and anomalies, as larger residuals\footnote{The word ``residuals'' is used to represent the specific residual values in residual features. For example, for a $C$ dimensional residual feature, it contains $C$ residual values (a.k.a, $C$ residuals).} are more likely to be anomalies than normal features. The normal features are usually closer to reference features compared to abnormal features, thus residuals of normal features will be closer to 0. To empirically validate this, we calculate the average absolute values of normal feature residuals and abnormal feature residuals. The calculated average absolute values are shown in Tab.\ref{tab:feature_decorrelation_statistics}.

Finally, we attempt to numerically evaluate the effectiveness of residual operation for feature decorrelation. Specifically, we calculate the Kurtosis of the multi-class feature distribution. For a multi-class distribution, Kurtosis can measure the variations of the distribution. When the class feature correlation is strong, features across classes will vary significantly, the multi-class feature distribution will be multi-modal, resulting in a small Kurtosis. When the class feature correlation is weak, the multi-class feature distribution tends to be single-modal, thereby the Kurtosis is large. The calculated Kurtosis values are shown in Tab.\ref{tab:feature_decorrelation_statistics}. The results validate that feature correlation can be effectively decreased in residual features, namely, residual operation can effectively eliminate class-related attributes in features.

\begin{table*}[ht]
\caption{\textbf{Statistics of the initial feature distribution and residual feature distribution}. ``Abs'' means the average absolute values of feature residuals. Anomalies are more likely to have larger residual values.}
\label{tab:feature_decorrelation_statistics}
\resizebox{0.7\linewidth}{!}{
\begin{tabular}{c|cccc}
\toprule[0.5mm]
  Dataset & MVTecAD & VisA & BTAD & MVTec3D \\
  \cmidrule(r){2-2} \cmidrule(r){3-3} \cmidrule(r){4-4} \cmidrule(r){5-5}
  Statistics & Kurtosis$\uparrow$ & Kurtosis$\uparrow$ & Kurtosis$\uparrow$ & Kurtosis$\uparrow$ \\
\midrule
 Initial Features & 0.261 & 0.757 & 0.190 & 0.047 \\

Residual Features & 1.679 & 1.313 & 0.518 & 0.552 \\
\cmidrule(r){2-2} \cmidrule(r){3-3} \cmidrule(r){4-4} \cmidrule(r){5-5}
Statistics & Abs & Abs & Abs & Abs \\
\midrule
 Normal Residual Features & 0.077 & 0.068 & 0.107 & 0.110 \\

Abnormal Residual Features & 0.392 & 0.365 & 0.335 & 0.409 \\

\bottomrule[0.5mm]
\end{tabular}}
\end{table*}

\subsection{Feature Hypersphere Constraining}
\label{sec:feature_constraintor}

\begin{figure}[ht]
    \centering
    \includegraphics[width=1.0\linewidth]{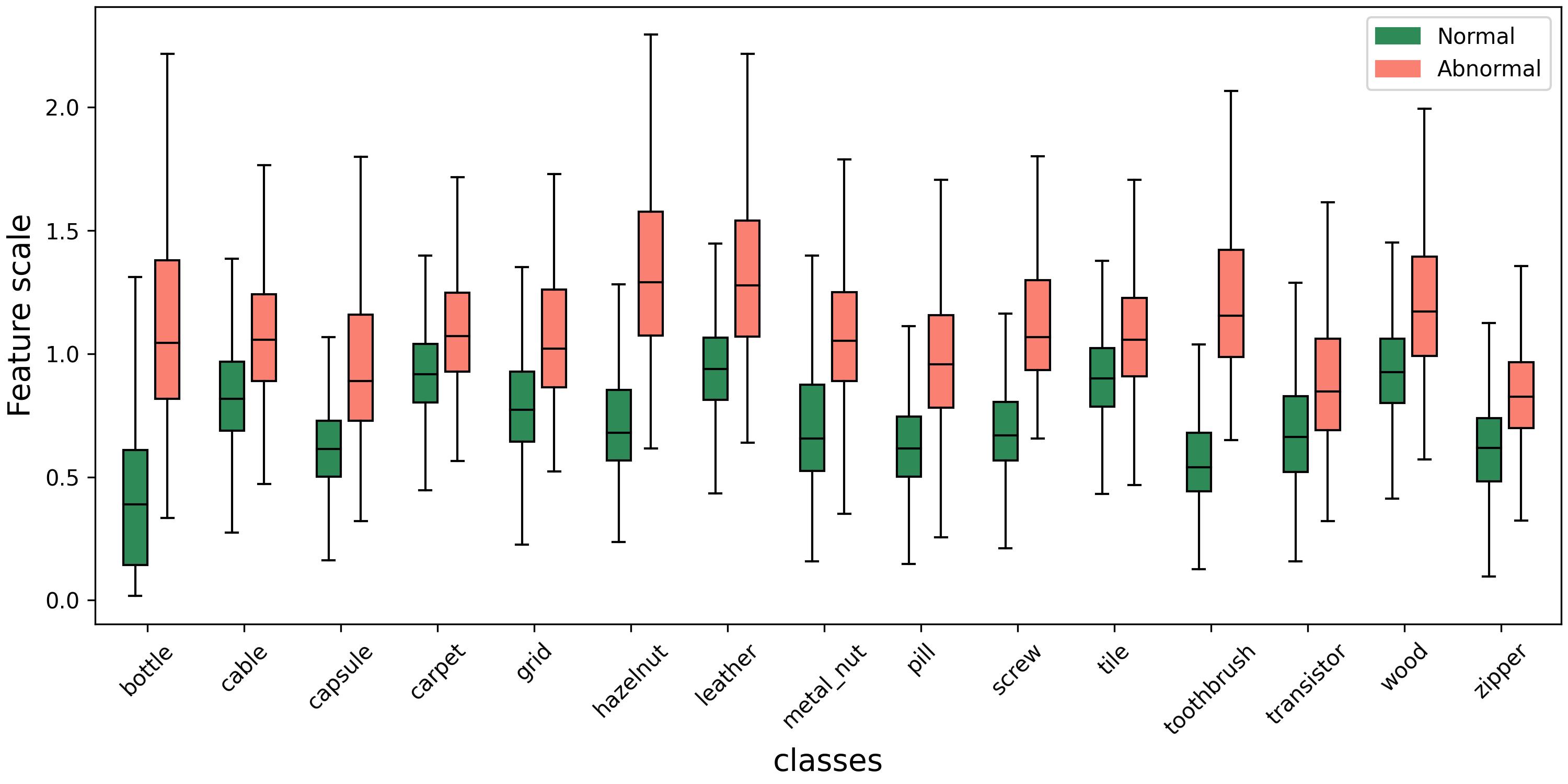}
    \caption{\textbf{Boxplot of feature scales of residual features on the MVTecAD dataset}. We utilize L2 norm as the statistical value of feature scale. From the perspective of feature scales to distinguish normals and anomalies, it can be seen that a good decision boundary in one class may be poor in another class.}
    \label{fig:vis_scale_correlation}
\end{figure}

Residual features can effectively realize feature decorrelation, but they still have one issue: class-related scale variation, \emph{a.k.a}, scale correlation. Scale correlation refers that features of different classes may still have significant differences in scale, namely, the numerical value scales in the features of different classes may be remarkably different. This can lead to difficulty in obtaining a unified normal-abnormal decision boundary for different classes, \emph{i.e.}, the scales of decision boundaries in different classes may be significantly different, a good decision boundary in one class may be poor in another class. To empirically show the differences in feature scales among different classes, we plot the boxplot of feature scales of those classes from the MVTecAD dataset in Fig.\ref{fig:vis_scale_correlation}. 

In order to further reduce variations in residual features and also maintain the consistency in feature scales among different classes (scale decorrelation), we take the idea from one-class-classification (OCC) learning \cite{deepSVDD, FCDD} and propose a \emph{Feature Constraintor} to constrain the initial normal residual features to a spatial hypersphere. The \emph{Feature Constraintor} $\psi(\cdot;\mathcal{W})$ projects the initial residual feature $x_i, i \in \{1,2,\dots,N\}$, where $N$\footnote{For symbol simplicity, we use $N$ to represent the number of normal residual features in one feature layer. Please note that our method is based on multi-layer features, with a \emph{Feature Constraintor} on each layer.} is the number of normal residual features, to the constrained feature $x_i^\prime$ as $x_i^\prime = \psi(x_i;\mathcal{W})$. This is the typical Hypersphere Contraction optimization problem \cite{SVDD}, where the goal is to seek a hypersphere that can enclose the transformed data around a given center. The general objective of this problem is formulated as follows:

\begin{equation}
\label{eq:general_objective}
    Dis(\psi(x_i;\mathcal{W}),\mathbf{c}) \le R, \quad i = 1,\dots,N
\end{equation}
where $\mathbf{c}$ is the center of the hypersphere, $R$ is the radius of the hypersphere, and $Dis(\cdot,\cdot)$ represents any distance function such as the $p$-norm distance, the cosine distance, and so on. Specifically, we employ pseudo-Huber loss \cite{Huber} to measure the distance. According to DeepSVDD \cite{SVDD}, a good way to set the center $\mathbf{c}$ is to initialize it to the mean of transformed features, \emph{i.e}, $\mathbf{c} = \frac{1}{N}\sum_{i=1}^{N}\psi(x_i;\mathcal{W})$. In our method, as the residual features are distributed around the origin, we simply fix the center as $0$, which is also easier to implement. With the center set to $0$, the distance is $Dis(\psi(x_i;\mathcal{W}),0) = \sqrt{||\psi(x_i;\mathcal{W})||_2^2+1} - 1$, which is a more robust distance measure that interpolates from quadratic to linear penalization \cite{Huber}. However, the inequality in Eq.(\ref{eq:general_objective}) is not a feasible optimization formula (\emph{i.e.}, no objectives to maximize or minimize). Then, we propose that the ideal optimization goal of Eq.(\ref{eq:general_objective}) is to make all features contracted into the hypersphere, and for features that are already within the hypersphere, there is no need to further shrink towards the center to avoid mode collapse. Then, the ideal optimization objective can be formulated as:
\begin{equation}
\label{eq:ideal_objective}
    \mathop{{\rm min}}\limits_{\mathcal{W}}\frac{1}{N}\sum_{i=1}^{N}I_{\_}(s_i) + \frac{\lambda}{2}\sum_{k=1}^{K}||\mathbf{W}_k||_F^2
\end{equation}
where $\lambda$ is the regularization parameter to reduce over-fitting ($\lambda$ is set to $0.001$), and $\mathbf{W}_k$ is the parameters of $k$th layer of the \emph{Feature Constraintor} with $\mathcal{W} = \{\mathbf{W}_1,\dots,\mathbf{W}_K\}$. And $s_i = D_i - R$ means the distance to the hypersphere, where $D_i$ refers to $Dis(\psi(x_i;\mathcal{W}),0)$, $I_{\_}(s_i)$ is the indicator function of non-positive real numbers, which is defined as:
\begin{equation}
    I_{\_}(s)=\left\{\begin{aligned}
        &0, \quad s \leq 0, \\
        &\infty, \quad s > 0.
    \end{aligned}\right.
\end{equation}
$I_{\_}(s)$ tends to infinity when the transformed feature $\psi(x_i;\mathcal{W})$ falls outside the hypersphere $(\mathbf{c}, R)$. Thus, minimizing Eq.(\ref{eq:ideal_objective}) can enforce the network to transform all features to fall inside the hypersphere $(\mathbf{c}, R)$. However, Eq.(\ref{eq:ideal_objective}) is difficult to directly optimize as the $I_{\_}(s)$ is not differentiable. To address this issue, we take insight from barrier methods \cite{ConvexOptimization} in convex optimization  and employ the logarithmic barrier function to smoothly approximate the indicator function $I_{\_}(s)$, that is
\begin{equation}
\label{eq:log_barrier_function}
    \hat{I}_{\_}(s) = -\frac{1}{t}{\rm log}(-s), \quad s < 0
\end{equation}
where $t$ controls the approximation precision of the logarithmic barrier function $\hat{I}_{\_}(s)$ to the indicator function $I_{\_}(s)$. Same as $I_{\_}(s)$, $\hat{I}_{\_}(s)$ is also convex and non-decreasing function. Especially, it increases to $\infty$ smoothly as $s$ increases to $0$, and according to the convention, we can take $\infty$ when $s \ge 0$. However, different from $I_{\_}(s)$, $\hat{I}_{\_}(s)$ is a differentiable closed function, moreover, the samples closer to the boundary of the hypersphere will obtain higher penalties as well as larger gradient values. In this way, we can pay more attention to the samples around the boundary, while those samples away from the boundary will not be further enforced to contract to the center, due to small losses and gradient values. With $t$ increasing, the approximation precision will gradually increase. However, in Eq.(\ref{eq:log_barrier_function}), when $s \geq 0$, the value of $\hat{I}_{\_}(s)$ is not well defined. But $s > 0$ indicates that the features are outside the boundary, we need to pay more attention to these features and contract them into the boundary. To this end, we further relax the limitation of $\hat{I}_{\_}(s)$ and propose the log-barrier OCC loss, which can be expressed as:
\begin{equation}
\label{eq:soft_log_barrier}
    \mathop{{\rm min}}\limits_{\mathcal{W}}-\frac{1}{Nt}\sum_{i=1}^{N}{\rm logsig}(-s_i)\cdot{\rm e}^{s_i} + \frac{\lambda}{2}\sum_{k=1}^{K}||\mathbf{W}_k||_F^2
\end{equation}
where ${\rm logsig}$ presents logarithmic sigmoid function. By employing the sigmoid function, we can ensure that $s$ can take values in the range $(-\infty, \infty)$. Then, we denote $J(\mathcal{W}) = -\frac{1}{Nt}\sum_{i=1}^{N}{\rm logsig}(-s_i)\cdot{\rm e}^{s_i}$ and $\sigma_i = {\rm sigmoid}(-s_i)$, the gradient of $J(\mathcal{W})$ with respect to $\mathcal{W}$ can be derived as:
\begin{align}
\label{eq:gradient}
    \frac{\partial J}{\partial \mathcal{W}} & = -\frac{1}{Nt}\sum_{i=1}^{N}\bigg(\frac{1}{\sigma_i}\frac{\partial \sigma_i}{\partial s_i}\frac{\partial s_i}{\partial \mathcal{W}}{\rm e}^{s_i} + {\rm log}\sigma_i{\rm e}^{s_i}\frac{\partial s_i}{\partial \mathcal{W}}\bigg) \nonumber \\
    & = \frac{1}{Nt}\sum_{i=1}^N\bigg(\frac{{\rm e}^{s_i}}{1+{\rm e}^{s_i}}\frac{\partial D_i}{\partial \mathcal{W}} - {\rm log}\sigma_i{\rm e}^{s_i}\frac{\partial D_i}{\partial \mathcal{W}}\bigg) \nonumber \\
    & = \frac{1}{Nt}\sum_{i=1}^N\bigg((1 - \sigma_i)\frac{\partial D_i}{\partial \mathcal{W}} - {\rm log}\sigma_i{\rm e}^{s_i}\frac{\partial D_i}{\partial \mathcal{W}}\bigg) 
\end{align}
According to the gradient formula in Eq.(\ref{eq:gradient}), we can better understand our log-barrier OCC loss in Eq.(\ref{eq:soft_log_barrier}), which can adaptively tune attention to different samples. The $\sigma_i$ can be explained as the probability that the feature belongs to the normal distribution. Thus, the features close to the center have large probabilities (\emph{i.e.}, $\sigma_i \rightarrow 1$) and thus small $1 - \sigma_i$ and $-{\rm log}\sigma_i$, while the features falling on the boundary and outside the boundary have small $\sigma_i \rightarrow 0$ and large $-{\rm log}\sigma_i$ (thus large gradient values). It is obvious that the objective in Eq.(\ref{eq:soft_log_barrier}) will assign larger gradients to the margin samples (\emph{i.e.}, close or outside the boundary) for better contraction. 

\begin{figure}[ht]
    \centering
    \includegraphics[width=1.0\linewidth]{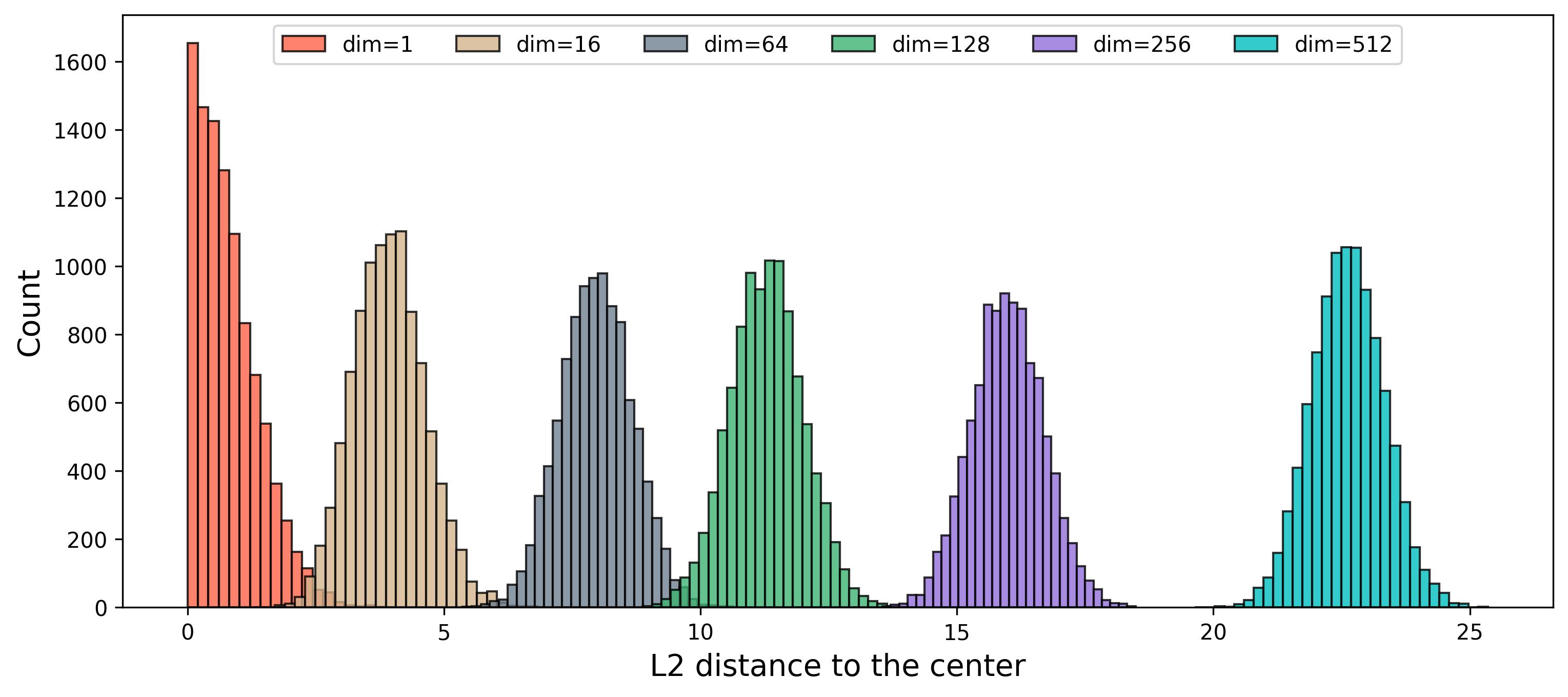}
    \caption{\textbf{\emph{Soap-bubble} phenomenon in the high-dimensional data}. All data are drawn from $\mathcal{N}(0, \textbf{I}_d)$, and we show the empirical histogram of distances to the center $0$ of $10^4$ samples.}
    \label{fig:soap_bubble}
\end{figure}

\textbf{Bidirectional Hypersphere Contraction.} Although the goal of Optimization (\ref{eq:soft_log_barrier}) is to promote most normal features close to the origin and distribute compactly around the origin. However, empirical exploration has found that in a high-dimensional space, most samples are far away from the origin \cite{SoupBubble}. This phenomenon is called \emph{soap-bubble} \cite{SoupBubble}, which means the high-dimensional data may be more likely to be located in the interval region of two hyperspheres instead of a hypersphere. In Fig.\ref{fig:soap_bubble}, we sample from Gaussian distribution $\mathcal{N}(0, \textbf{I}_d)$ and show the histogram of distances to the center. It can be seen that the Gaussian is in the thin shell within a distance from the origin when the dimension is large. Moreover, the higher the dimensionality of the data, the more sampled instances are far from the center. The \emph{soap-bubble} phenomenon is formally proven by the following proposition (\emph{cf.} Lemma 1 of \cite{HD-Theory}).

\textbf{Proposition 1.} \emph{Suppose $z_1,z_2,\dots,z_n$ are sampled from $\mathcal{N}(0, \textbf{I}_d)$ independently. Then, for any $z_i$ and all $t \geq 0$, the following inequality hold.}
\begin{equation}
    {\rm P}\Big[||z_i|| \geq \sqrt{d-2\sqrt{dt}}\Big] \geq 1 - e^{-t}
\end{equation}
The proposition shows that when the dimension is high, each $z_i$ is outside the hypersphere of radius $R^\prime := \sqrt{d - 2\sqrt{dt}}$ with a probability of at least $1 - {\rm e}^{-t}$. When $R^\prime$ is closer to $R$, normal data are more likely to be away from the center. For example, when $R$ is 1 and $d$ is 256, the proposition gives the probability that $z_i$ is outside the hypersphere of radius $R$ is almost 1.

Therefore, to avoid the sparsity in the hypersphere of the normal data distribution and narrow the scope of the decision area, the optimization objective should obey the bi-hypersphere property of high-dimensional data (see Fig.\ref{fig:soap_bubble}). Inspired by \cite{DO2HSC}, we aim to compress normal data to an interval region between two co-centered hyperspheres. Combined with our log-barrier OCC loss in Eq.(\ref{eq:soft_log_barrier}), we propose the following log-barrier bi-contraction OCC loss:
\begin{align}
\label{eq:bi_hypersphere_log_barrier}
    \mathcal{L}_{bi-occ} = -\frac{1}{Nt}&\sum_{i=1}^{N}\big({\rm logsig}(R_{{\rm max}} - D_i)\cdot{\rm e}^{D_i - R_{{\rm max}}} \nonumber \\
    & + {\rm logsig}(D_i - R_{{\rm min}})\cdot{\rm e}^{R_{{\rm min}} - D_i}\big)
\end{align}
where we replace $s_i$ with $D_i - R_{{\rm max}}$ and $R_{{\rm min}} - D_i$. From the previous analysis of Eq.(\ref{eq:soft_log_barrier}), we can know that when $D_i > R_{{\rm max}}$ and $D_i < R_{{\rm min}}$, the loss in Eq.(\ref{eq:bi_hypersphere_log_barrier}) will generate large gradients to compress the features outside the boundaries into the bi-hypersphere interval region.



\textbf{Anomaly-invariant Constraint Item.} The log-barrier bi-contraction OCC loss in Eq.(\ref{eq:bi_hypersphere_log_barrier}) can be used for constraining the normal residual features to a bi-hypersphere interval region. However, if we only constrain features to the interval region, the network may more easily overfit and simply map all features to the region. If we give the network another objective for anomalous features, this will urge the network to distinguish between normal and abnormal, rather than forming a shortcut solution. Thus, we further introduce an anomaly-invariant constraint item by simply predicting the initial features to form an anomaly-invariant OCC loss to optimize our \emph{Feature Constraintor}. The loss is defined as:
\begin{equation}
    \mathcal{L}_{ai-occ} = \mathcal{L}_{bi-occ} + \frac{1}{M}\sum_{j=1}^{M}||x_{j}^{\prime} - x_{j}||_2
\end{equation}
where $M$ is the number of abnormal residual features, $x_j,j\in\{1,2,\dots,M\}$ represents each abnormal residual feature, and $x_j^\prime = \psi(x_j;\mathcal{W})$. ``Invariant'' means the abnormal residual features remain relatively unchanged to themselves and will not be mapped into the hypersphere. In this way, our proposed anomaly-invariant OCC loss can not only make the distribution of normal residual features more compact but also keep abnormal residual features as invariant as possible. In addition, by constraining normal features into a bi-hypersphere interval region, the normal feature scales of different classes can also be more consistent. Therefore, after the \emph{Feature Constraintor}, the normal and abnormal residual features are more distinguishable (see Fig.\ref{fig:vis_scale_decorrelation}), namely, we can obtain a better unified decision boundary.

Finally, we also numerically evaluate the effectiveness of feature hypersphere constraining for scale decorrelation. Specifically, we utilize L2 norm as the statistical value of feature scale. For each class, we calculate an average feature scale value. Then, we calculate the standard deviation of feature scale values of multiple classes. When the scale correlation is strong, scales of features across classes will vary significantly, resulting in a large standard deviation. When the scale correlation is weak, scales of features tend to be consistent, thereby the standard deviation is small. The numerical statistical values are shown in Tab.\ref{tab:scale_decorrelation_statistics}.

\begin{table*}[ht]
\caption{\textbf{Statistics of initial residual features and constrained residual features.} ``Std'' means the standard deviation.}
\label{tab:scale_decorrelation_statistics}
\resizebox{0.7\linewidth}{!}{
\begin{tabular}{c|cccc}
\toprule[0.5mm]
  Dataset & MVTecAD & VisA & BTAD & MVTec3D \\
  \cmidrule(r){2-2} \cmidrule(r){3-3} \cmidrule(r){4-4} \cmidrule(r){5-5}
  Statistics & Std$\downarrow$ & Std$\downarrow$ & Std$\downarrow$ & Std$\downarrow$ \\
\midrule
 Initial Residual Features & 0.147 & 0.098 & 0.146 & 0.059 \\

Constrained Residual Features & 0.005 & 0.004 & 0.004 & 0.002 \\
\bottomrule[0.5mm]
\end{tabular}}
\end{table*}

\subsection{VQ-based Feature Distribution Matching}
\label{sec:fdm}

Residual features and feature hypersphere constraining have been able to effectively reduce the feature variations in new classes. In this subsection, we aim to further enhance the consistency of testing and training data distributions. To this end, we propose a vector quantization (VQ) based feature distribution matching approach, which can contribute to improving the model's class-adaptive ability on novel classes. Feature Distribution Matching (FDM) is a group of techniques that aims to reduce the distribution mismatch or discrepancy of data from two different domains \cite{MixStyle, EFDM}. To ensure the generalization capability of our AD model in novel classes, we also want the testing residual features can obey the training feature distribution as much as possible. This also conforms to the common sense in machine learning, where the more consistent the testing data distribution and the training data distribution, usually the better the model's performance. Therefore, we think that FDM will be a promising and effective technique for class-agnostic anomaly detection.

Specifically, following \cite{ADShift}, we adopt the EFDM \cite{EFDM} to inject training distribution into the testing residual features, which is the SOTA of FDM. EFDM can precisely match empirical Cumulative Distribution Functions of image features, resulting in exact feature distribution alignment and accurate matching of statistical properties like mean, standard deviation, and high-order statistics. Formally, for residual features $\mathcal{Q} \in \mathbb{R}^{N \times C}$ of a test sample, we need to select the same number of residual features from the training distribution, denoted as $\mathcal{P} \in \mathbb{R}^{N \times C}$. Then, the EFDM is carried out as follows:
\begin{equation}
\label{eq:fdm}
    {\rm EFDM}(\mathcal{Q}, \mathcal{P}, \alpha) : \mathcal{Q}_{\tau_i} = \alpha\mathcal{Q}_{\tau_i} + (1-\alpha) \mathcal{P}_{\omega_i}
\end{equation}
where $\{\mathcal{Q}_{\tau_i}\}_{i=1}^{NC}$ and $\{\mathcal{P}_{\omega_i}\}_{i=1}^{NC}$ are sorted values of $\mathcal{Q}$ and $\mathcal{P}$ in ascending order. Here, $NC$ represents the number of elements in $\mathcal{Q}$ and $\mathcal{P}$. From the perspective of style transfer \cite{StyleTransfer1, StyleTransfer2}, $\mathcal{Q}$ can be understood as playing the role of carrying the appearance information, and $\mathcal{P}$ plays the role of carrying the style information. In our method, the $\mathcal{P}$ plays the role in conveying distribution information pertaining to the training features. However, the training features are tremendous, it's not practical to represent the distribution by storing all training features, and this can also lead to feature selection being very computationally expensive. To this end, we employ the vector quantization \cite{VQ-VAE} technique during training to learn a codebook $\mathcal{C} \in \mathbb{R}^{K \times C}$ that contains $K$ discrete codebook embeddings to represent training data. Based on PyTorch, we can simply use \emph{nn.Embedding(K, C)} to construct the codebook. For a feature $x_i$ passed into the vector quantization module, the nearest element from the codebook $\mathcal{C}$ will be used as the quantized feature $x_q$ for $x_i$. Then, the vector quantization loss is defined as follows:
\begin{equation}
    \mathcal{L}_{vq} = \frac{1}{N}\sum_{i=1}^N||sg[x_q] - x_i||_2^2 + \beta||x_q - sg[x_i]||_2^2
\end{equation}
where $sg[\cdot]$ stands for the stop-gradient operator and $\beta$ is set to 0.25. The advantage of the codebook is that it not only significantly reduces the number of stored features but also can effectively represent the training feature distribution \cite{VQ-VAE, VQ2}. Then, with the learned codebook $\mathcal{C}$, we can look up the nearest neighbor in the codebook for each feature in $\mathcal{Q}$ and thus obtain $\mathcal{P}$. Finally, we note that in implementation, we construct a specific codebook at each feature layer, and FDM is also performed at each layer.

\subsection{Feature Distribution Estimating}

We employ the normalizing flow (NF) model \cite{realNVP} as our \emph{Feature Distribution Estimator} to estimate the residual feature distribution. Note that our framework is not limited to normalizing flow, and other probabilistic models can also be used as the distribution estimator. Formally, we denote $\varphi(\cdot;\theta) : \mathcal{X} \in \mathbb{R}^{C} \rightarrow \mathcal{Z} \in \mathbb{R}^{C}$ as our normalizing flow. The input residual feature $x_{i}^{\prime}$ will be transformed into a latent feature $z_{i} = \varphi(x_{i}^{\prime};\theta)$ by the NF model. The estimated residual distribution $p_{\theta}(x)$ can be calculated according to the change of variables formula as follows \cite{realNVP, Glow}:
\begin{equation}
    {\rm log}p_{\theta}(x) = {\rm log}p_Z(z) + {\rm log}|{\rm det}J|
\end{equation}
where the $J = \nabla_xz$ is the Jacobian matrix of the bijective transformation $\varphi(\cdot;\theta)$. The model parameters $\theta$ can be optimized by maximizing the log-likelihoods and the latent variable $Z$ is composed of $Z_n$ and $Z_a$. $Z_n$ is the normal base distribution for normal features and can be assumed to obey $\mathcal{N}(0, \mathbf{I})$. To distinguish normal and abnormal features, the latent base distribution for abnormal features needs to have a certain distance from the normal base distribution $Z_n$, we denote it as $Z_a = \mathcal{N}(a, \mathbf{I})$, where $a$ is a hyperparameter to control the distance between two base distributions. We set $a=1$ in our model. Then, the log-likelihood of each feature $x_i^\prime$ corresponding to the normal and abnormal base distribution can be derived as follows, respectively
\begin{equation}
    {\rm log}p_{\theta}^{Z_n}(x_i^\prime) = -\frac{C}{2}{\rm log}(2\pi) + \frac{1}{2}||z_i||_2^2 - {\rm log}|{\rm det}J_i| \nonumber 
\end{equation}
and,
\begin{equation}
    {\rm log}p_{\theta}^{Z_a}(x_i^\prime) = -\frac{C}{2}{\rm log}(2\pi) + \frac{1}{2}||z_i - a||_2^2 - {\rm log}|{\rm det}J_i|
\end{equation}

The maximum likelihood loss (minimize negative log-likelihoods) for optimizing the NF model to fit the normal and abnormal residual features is derived as:
\begin{align}
\label{eq:loss_ml}
    \mathcal{L}_{ml} = \frac{-1}{N+M}\sum_{i=1}^{N+M}(1 - y_i){\rm log}p_{\theta}^{Z_n}(x_i^\prime) + y_i{\rm log}p_{\theta}^{Z_a}(x_i^\prime)
\end{align}
where the binary indicator $y_i$ is set to be $1$ when the corresponding position is anomalous and $y_i = 0$ denotes that the corresponding location is normal. In practice, we can downsample the ground-truth mask to the size of each feature map, which can indicate normal and abnormal positions. 

Furthermore, we construct the anomaly classification score $s(x_i^\prime) = \frac{p_{\theta}^{Z_a}(x_i^\prime)}{p_{\theta}^{Z_n}(x_i^\prime) + p_{\theta}^{Z_a}(x_i^\prime)}$ (\emph{i.e.}, $s(x_i^\prime)$ measures the probability of $x_i^\prime$ being classified as abnormal). Accordingly, we can also construct the normal classification
score: $1 - s(x_i^\prime)$. Then, apart from the basic maximum likelihood loss, based on our constructed normal/anomaly classification scores, we can also use binary classification loss for optimization. Naturally, considering the imbalance between normal and abnormal, we employ the focal loss \cite{FocalLoss} as follows:
\begin{equation}
\label{eq:loss_focal}
    \mathcal{L}_{focal} = \frac{1}{N+M}\sum_{i=1}^{N+M}\mathbf{FOCAL}(s(x_i^\prime), y_i)
\end{equation}
where $\mathbf{FOCAL}$ denotes the focal loss. The Eq.(\ref{eq:loss_ml}) and Eq.(\ref{eq:loss_focal}) both allow the NF model to transform normal and abnormal features to two distinct base distributions, thereby enhancing the discriminability. 


Then, the whole loss function for training the NF model is as follows:
\begin{equation}
    \mathcal{L}_{nf} = \mathcal{L}_{ml} + \mathcal{L}_{focal}
\end{equation}


\subsection{Inference and Anomaly Scoring}
\label{sec:anomaly_scoring}
 For new classes, our method only requires few-shot normal samples to extract some features as reference, without any fine-tuning. We feed each test feature $x^l_i$ from layer $l$ into the \emph{Feature Constraintor} $\psi(\cdot;\mathcal{W})$ and the \emph{Feature Distribution Estimator} $\varphi(\cdot;\theta)$ to get the latent feature $z^l_i$. In anomaly detection, anomalies are treated as outliers. Thus, the anomaly score is calculated based on the normal base distribution $Z_n$:
\begin{equation}
    s(x^l_i) = 1 - {\rm exp}\bigg(-\frac{C_l}{2}{\rm log}(2\pi) - \frac{1}{2}||z_i^l||_2^2 + {\rm log}|{\rm det}J^l_i|\bigg)
\end{equation}
Then, we upsample all $s(x^l_i)$ in the $l$th layer to the input image resolution ($H_0\times W_0$) using bilinear interpolation and combine all layers to obtain the anomaly score map. In addition, we also use the anomaly classification score $s(x_i^l) = \frac{p_{\theta}^{Z_a}(x_i^{\prime,l})}{p_{\theta}^{Z_n}(x_i^{\prime,l}) + p_{\theta}^{Z_a}(x_i^{\prime,l})}$ as anomaly score, and upsample all scores to the input image resolution. Then, the final anomaly score map is obtained by averaging the two above anomaly score maps. The maximum score of the anomaly score map is taken as the anomaly detection score of the image.

\section{Experiments}
\subsection{Datasets}

\textbf{Datasets.} We conduct extensive experiments on 8 real-world datasets, covering various industrial inspection scenarios and medical imaging domains, to evaluate the performance of our ResAD++ sufficiently. In industrial inspection, we consider the popular benchmarks, including MVTecAD \cite{MVTec}, VisA \cite{VisA}, BTAD \cite{BTAD}, MVTec3D \cite{MVTec3D}, and MPDD \cite{MPDD}. We also utilize the MVTecLOCO \cite{MVTecLOCO} dataset for logical anomaly detection. In medical imaging, we consider colon polyp detection dataset Kvasir-Seg \cite{Kvasir-SEG} and brain tumor detection dataset BraTS \cite{BRATS}.

\textbf{MVTecAD.} The MVTecAD \cite{MVTec} dataset is widely used as a standard benchmark for evaluating unsupervised image anomaly detection methods. This dataset contains 5354 high-resolution images of 15 different product categories, in which 5 classes consist of textures and the other 10 classes contain objects. It comprises 3692 defect-free training samples, as well as 1725 test images with and without defects.

\textbf{VisA.} The VisA \cite{VisA} dataset is another widely used unsupervised anomaly detection dataset. This dataset contains 10821 images with 9621 normal and 1200 anomalous samples. In addition to images that only contain a single instance, the VisA dataset also has images that contain multiple instances. Moreover, some product categories of this dataset, such as Cashew, Chewing gum, Fryum, and Pipe fryum, have objects that are roughly aligned. These characteristics make the VisA dataset more challenging than the MVTecAD dataset.

\textbf{BTAD.} The BeanTech Anomaly Detection dataset \cite{BTAD} contains 2830 images of 3 industrial products. The dataset is mainly used for texture anomaly detection, with all three products containing rich textures. Product 1, 2, and 3 of this dataset contain 400, 1000, and 399 training images, respectively.

\textbf{MVTec3D.} The MVTec3D \cite{MVTec3D} dataset is proposed for 3D anomaly detection, which contains 4147 3D point cloud scans paired with 2D RGB images from 10 real-world categories. In this dataset, most anomalies can also be detected only through RGB images. Since we focus on image anomaly detection, we only use RGB images of the MVTec3D dataset.

\textbf{MPDD.} MPDD \cite{MPDD} is focused specifically on defect detection during painted metal part fabrication, containing 6 classes of metal parts. This dataset poses more challenges than previous AD datasets, \emph{i.e.}, images are captured under complex conditions, including variable spatial orientations, positions, and distances of multiple objects concerning the camera at different light intensities and with a non-homogeneous background.

\textbf{MVTecLOCO.} The MVTecLOCO \cite{MVTecLOCO} dataset is specifically proposed for logical anomaly detection. It contains five object categories from industrial inspection scenarios with a total of 1772 images for training and 1568 images for testing. The anomalies in MVTecLOCO mainly manifest themselves in the violation of logical constraints, \emph{e.g.}, the screw bag contains two long screws and lacks a short one.


\textbf{BraTS.} BraTS \cite{BRATS} is a multimodal magnetic resonance imaging (MRI) dataset for brain tumor segmentation, which is from the famous BraTS Challenge. We use the training set from the 2021 BraTS Challenge in this paper. The dataset contains 1251 brain MRIs acquired by 19 institutions employing different clinical protocols. Each brain MRI comprises multiple manually annotated MRI scans with the shape of $240\times 240 \times 155$ ($H\times W \times D$). To evaluate image AD models, we extract the 83rd MRI scan from each brain MRI and convert them to images. We can finally obtain 1097 images that contain brain tumors and 154 tumor-free images for testing.

\textbf{Kvasir-Seg.} The Kvasir-Seg \cite{Kvasir-SEG} dataset is based on the Kvasir \cite{Kvasir} dataset, which is a multi-class dataset for gastrointestinal (GI) tract disease detection and classification. The original Kvasir dataset comprises 8000 GI tract images from 8 classes. However, the dataset is limited to classification tasks, due to only image-wise annotations. The Kvasir-Seg dataset is based on the polyp class of the Kavsir dataset, and the polyp images are manually annotated by a medical doctor and then verified by an experienced gastroenterologist.


\begin{table*}[ht]
\caption{\textbf{Anomaly detection and localization results with AUROC metric (\%) on eight real-world AD datasets under various few-shot AD settings}. $\cdot/\cdot$ means image-level and pixel-level AUROCs. In ResAD$^{\dag}$++, we utilize the ImageBind \cite{ImageBind} as the feature extractor. RDAD and UniAD don't utilize the few-shot normal samples to fine-tune. AnomalyCLIP is a zero-shot AD model as a performance baseline in new classes, so the results under 2-shot, 4-shot, and 8-shot are the same. The best results are in \textbf{bold}, and the second best are \underline{underlined}.}
\label{tab:main_results}
\resizebox{1.0\linewidth}{!}{
\begin{tabular}{c|c||cc|cccc|c|ccc|ca}
\toprule[0.5mm]
  \multirow{3}*{\textbf{Setting}} & \multirow{3}*{\textbf{Datasets}} & \multicolumn{2}{c|}{\textbf{Baselines}} & \multicolumn{4}{c|}{\textbf{Few-shot AD Methods (Non-CLIP-based)}} & \multirow{3}*{\makecell{\textbf{ResAD++} \\ \textbf{(W50)}}} & \multicolumn{3}{c|}{\textbf{CLIP-based AD Methods}} & \multirow{3}*{\makecell{ResAD$^{\dag}$ \\ NeurIPS2024}} &  \\
   &  & \makecell{RDAD \\ CVPR2022}& \makecell{UniAD \\ NeurIPS2022} & SPADE & PaDiM & \makecell{PatchCore \\ CVPR2022} & \makecell{RegAD \\ ECCV2022} &  & \makecell{WinCLIP \\ CVPR2023} & \makecell{AnomalyCLIP \\ ICLR2024} & \makecell{InCTRL \\ CVPR2024} & & \multirow{-2}*{\makecell{\textbf{ResAD$^{\dag}$++}}}\\
\midrule
 \multirow{9}*{\textbf{2-shot}} & MVTecAD & 70.4/77.3 & 68.2/80.5 & 74.6/92.6 & 79.0/93.2 & 83.2/92.1 & 80.4/93.3 & 87.8/\underline{94.6} & 93.1/93.8 & 91.5/91.1 &\underline{94.0}/- & 94.4/95.6 & \textbf{96.0}/\textbf{95.9} \\
 
    & VisA & 61.3/80.9 & 49.9/79.7 & 71.7/88.7 & 62.8/90.5 & 77.4/91.3 & 69.4/93.3 & \underline{87.3}/\underline{96.5} & 81.9/94.9 & 82.1/95.5 & 85.8/- & 84.5/95.1 & \textbf{90.8}/\textbf{97.6}  \\
  
   & BTAD & 83.0/92.9 & 69.7/82.4 & 80.7/92.5 & 88.9/95.9 & 86.5/92.8 & 87.2/93.9 & \textbf{93.1}/\underline{96.4} & 87.5/95.8 & 88.3/94.2 & 92.3/- & 91.1/96.4 & \underline{92.9}/\textbf{97.3}  \\
  
   & MVTec3D & 55.8/93.5 & 52.2/89.1 & 62.5/93.6 & 56.5/95.4 & 62.9/94.5  & 59.5/95.4 & 69.4/\underline{97.1} & \underline{72.3}/96.8 & 65.8/94.8 & 68.9/- & 78.5/97.5 & \textbf{82.9}/\textbf{97.4} \\

   & MPDD & 59.8/75.9 & 54.3/86.9 & 57.3/96.0 & 52.9/90.6 & 55.2/90.0  & 54.1/93.9 & 70.4/96.4 & 75.3/94.5 & \underline{77.0}/\underline{96.5} & 71.5/- & 77.3/97.3 & \textbf{78.2}/\textbf{97.5} \\

   & MVTecLOCO & 47.3/54.8 & 49.1/51.1 & 59.3/65.7 & 53.2/66.4 & 57.5/\underline{68.0}  & 55.0/66.2 & \underline{64.5}/67.4 & 62.6/64.7 & 55.6/63.5 & 58.7/- & 65.6/68.0 & \textbf{66.3}/\textbf{68.5} \\

   & BraTS & 49.8/66.7 & 59.5/88.5 & 58.0/92.8 & 49.4/90.2 & 58.2/\underline{93.5}  & 54.6/81.4 & 66.1/91.4 & 55.9/91.5 & 63.4/90.8 & \textbf{74.6}/- & 67.9/94.3 & \underline{69.0}/\textbf{94.7} \\

   & Kvasir-Seg & 54.4/54.2 & 50.3/49.9 & 85.7/67.8 & 86.8/65.4 & 79.0/69.0  & 70.1/51.0 & \textbf{86.5}/69.2 & 76.2/73.0 & 75.9/\underline{78.9} & 63.3/- & 77.5/79.4 & \underline{80.5}/\textbf{80.3} \\

    & \textbf{Average} & 60.2/74.5 & 56.7/76.0 & 68.7/86.2 & 66.2/86.0 & 70.0/86.4 & 66.3/83.6 & \underline{78.1}/\underline{88.6} & 75.6/88.1 & 75.0/88.2 & 76.1/- & 79.6/90.4 & \textbf{82.1}/\textbf{91.2} \\

 \midrule
 
 \multirow{9}*{\textbf{4-shot}} & MVTecAD & 70.4/77.3 & 68.2/80.5 & 75.5/94.3 & 82.0/94.4 & 87.4/94.2 & 84.8/94.5 & 90.8/\underline{95.8} & \underline{94.6}/94.2 & 91.5/91.1 & 94.5/- & 94.2/96.9 & \textbf{96.4}/\textbf{96.3} \\
 
    & VisA & 61.3/80.9 & 49.9/79.7 & 75.0/93.7 & 70.3/93.4 & 82.6/94.0 & 78.0/93.5 & \underline{89.3}/\underline{96.8} & 84.1/95.2 & 82.1/95.5 & 87.7/- & 90.8/97.5 & \textbf{92.1}/\textbf{98.0} \\
  
   & BTAD & 83.0/92.9 & 69.7/82.4 & 84.0/95.2 & \underline{91.9}/96.9 & 91.1/95.5 & 90.8/94.9 & \textbf{94.1}/\textbf{97.3} & 89.5/95.8 & 88.3/94.2 & 91.7/- & 91.5/96.8 & 91.0/\underline{97.2} \\
  
   & MVTec3D & 55.8/93.5 & 52.2/89.1 & 61.6/95.0 & 60.0/95.8 & 65.3/95.5 & 62.3/96.7 & 71.1/\underline{97.6} & \underline{74.1}/97.0 & 65.8/94.8 & 69.1/- & 82.4/97.9 & \textbf{83.5}/\textbf{97.7} \\

    & MPDD & 59.8/75.9 & 54.3/86.9 & 62.4/97.0 & 52.9/92.8 & 67.9/95.5 & 66.2/94.1 & 75.6/\underline{97.4} & \underline{79.4}/95.1 & 77.0/96.5 & 71.7/- & 86.0/98.0 & \textbf{86.5}/\textbf{98.1} \\

    & MVTecLOCO & 47.3/54.8 & 49.1/51.1 & 64.0/66.8 & 54.7/67.9 & 61.7/\underline{69.4} & 56.6/66.1 & \underline{67.1}/67.3 & 65.5/65.2 & 55.6/63.5 & 60.8/- & 70.0/69.0 & \textbf{70.8}/\textbf{69.5} \\
    
   & BraTS & 49.8/66.7 & 59.5/88.5 & 66.3/94.8 & 60.6/94.5 & 71.2/\underline{95.9} & 60.0/87.3 & 74.9/94.2 & 67.3/93.2 & 63.4/90.8 & \underline{76.9}/- & 84.6/96.1 & \textbf{85.7}/\textbf{96.4} \\

   & Kvasir-Seg & 54.4/54.2 & 50.3/49.9 & \textbf{89.6}/67.9 & 87.1/65.8 & 84.0/69.0 & 72.7/53.6 & \underline{89.0}/70.2 & 78.3/73.0 & 75.9/\underline{78.9} & 65.0/- & 79.1/80.4 & 81.2/\textbf{81.2} \\
  
    & \textbf{Average} & 60.2/74.5 & 56.7/76.0 & 72.3/88.1 & 69.9/87.7 & 76.4/88.7 & 71.4/85.1 & \underline{81.5}/\underline{89.6} & 79.1/88.6 & 75.0/88.2 & 77.2/- & 84.8/91.6 & \textbf{85.9}/\textbf{91.8} \\

\midrule

\multirow{9}*{\textbf{8-shot}} & MVTecAD & 70.4/77.3 & 68.2/80.5 & 78.9/95.7 & 85.0/95.6 & 90.2/94.8 & 88.2/95.9 & 93.1/\underline{96.4} & 94.8/94.5 & 91.5/99.1 & \underline{95.3}/- & 97.7/96.7 & \textbf{98.6}/\textbf{96.7} \\
 
    & VisA & 61.3/80.9 & 49.9/79.7 & 73.7/94.2 & 77.9/95.0 & 85.4/93.9 & 80.0/95.3 & \underline{90.6}/\underline{97.2} & 85.4/95.4 & 82.1/95.5 & 88.7/- & 92.3/97.8 & \textbf{93.4}/\textbf{98.2} \\
  
   & BTAD & 83.0/92.9 & 69.7/82.4 & 84.2/\underline{96.3} & \underline{93.2}/97.2 & 91.4/96.0 & 91.6/\textbf{97.3} & \textbf{94.0}/\textbf{97.3} & 90.2/96.0 & 88.3/94.2 & 89.0/- & 91.6/96.8 & 91.9/\textbf{97.3} \\
  
   & MVTec3D & 55.8/93.5 & 52.2/89.1 & 63.8/95.6 & 63.5/96.3 & 68.3/94.2 & 67.4/96.9 & 74.5/\textbf{98.0} & \underline{75.8}/97.1 & 65.8/94.8 & 71.4/- & 83.0/97.9 & \textbf{85.2}/\underline{97.8} \\

    & MPDD & 59.8/75.9 & 54.3/86.9 & 62.3/\underline{97.3} & 56.3/94.0 & 70.3/95.1 & 74.7/95.6 & 80.0/\textbf{98.0} & \underline{81.7}/94.7 & 77.0/96.5 & 75.8/- & 87.9/97.8 & \textbf{88.3}/\textbf{98.0} \\

    & MVTecLOCO & 47.3/54.8 & 49.1/51.1 & 64.3/67.3 & 57.7/69.1 & 65.7/\underline{69.7} & 62.2/69.6 & 67.8/67.6 & \underline{68.0}/65.5 & 55.6/63.5 & 62.9/- & 69.5/69.3 & \textbf{69.8}/\textbf{69.9} \\
    
   & BraTS & 49.8/66.7 & 59.5/88.5 & 72.6/95.4 & 71.2/96.0 & 76.4/\underline{96.4} & 66.6/86.1 & 79.1/95.1 & 68.9/93.8 & 63.4/90.8 & \underline{79.3}/- & 85.9/96.3 & \textbf{87.9}/\textbf{96.8} \\

   & Kvasir-Seg & 54.4/54.2 & 50.3/49.9 & 90.1/68.7 & \textbf{91.2}/67.1 & 84.5/68.4 & 76.7/51.6 & \underline{90.5}/68.8 & 83.1/73.1 & 75.9/\underline{78.9} & 75.0/- & 81.5/81.2 & 83.4/\textbf{82.1} \\
  
    & \textbf{Average} & 60.2/74.5 & 56.7/76.0 & 73.7/88.8 & 74.5/88.8 & 79.0/88.6 & 75.9/86.0 & \underline{83.7}/\underline{89.8} & 81.0/88.8 & 75.0/88.2 & 79.7/- & 86.2/91.7 & \textbf{87.3}/\textbf{92.1} \\
  
\bottomrule[0.5mm]
\end{tabular}}
\end{table*}

\subsection{Evaluation Metrics}

For both image-level anomaly detection and pixel-level anomaly localization, the standard metric in anomaly detection, the area under the receiver operating characteristics curve (AUROC), will be used to evaluate the performance of AD methods. Nonetheless, as abnormal areas in the image are usually smaller than the normal areas, this may cause overestimated pixel-level AUROC values. This means that for some small anomalies, even if they are not correctly located, the pixel-level AUROC is still high. Thus, to more accurately measure the performance of anomaly localization, we also adopt the Per-Region-Overlap (PRO) curve metric proposed in \cite{STAD}. The PRO score can take into account the overlap and recovery of connected anomaly components to better account for varying anomaly sizes, see \cite{STAD} for details. Specifically, we report the PRO score with 0.3 FPR, which means that the normalized area under the PRO curve up to an average false positive rate per-pixel of 30\%.


\subsection{Implementation Details}

\textbf{Implementation Details.} All the training and test images are resized and cropped to $224 \times 224$ resolution. The parameters of the feature extractor are frozen during training. The layer numbers of the NF models are all 10. We use the Adam \cite{Adam} optimizer with weight decay $5e^{-4}$ to train the model. The total training epochs are set to 100, and the batch size is 32 by default. The learning rate is $1e^{-5}$ initially and dropped by 0.1 after $[70, 90]$ epochs. The boundary $R_{max}$ is dynamic, $R_{max} = {\rm min}({\rm max}(\{Dis(x_j, 0)\}_{j=1}^{M}), 0.4)$, where $Dis(x_j, 0)$ is the distance from abnormal feature $x_j$ to origin. $R_{min}$ is set as $0.99*R_{max}$. The number of discrete quantization vectors in the VQ module is set to 1536, and the $\alpha$ in the FDM module is set to 0.4.

We follow CFLOW \cite{CFLOW} to implement the normalizing flow model. The normalizing flow model is mainly based on Real-NVP \cite{realNVP} architecture, which is composed of the so-called coupling layers. All coupling layers have the same architecture, where a learnable subnet is utilized to predict the affine parameters \cite{realNVP}. The convolutional subnet in Real-NVP is replaced with a two-layer MLP network. Each coupling layer is followed by a random and fixed soft permutation of channels \cite{SoftPerm} and a fixed scaling by a constant, similar to ActNorm layers introduced by \cite{Glow}. Furthermore, we adopt the soft clamping of multiplication coefficients used by \cite{realNVP}, the clamping coefficient is set to 1.9.


We run all the experiments with the NVIDIA RTX 4090 GPU and random seed 42. 


\begin{table*}[ht]
\caption{\textbf{Anomaly localization results with PRO metric (\%) on eight real-world AD datasets under various few-shot AD settings}. The InCTRL only provides image-level anomaly detection results, so the PRO scores of InCTRL are missing.}
\label{tab:main_results2}
\resizebox{1.0\linewidth}{!}{
\begin{tabular}{c|c||cc|cccc|c|ccc|ca}
\toprule[0.5mm]
  \multirow{3}*{\textbf{Setting}} & \multirow{3}*{\textbf{Datasets}} & \multicolumn{2}{c|}{\textbf{Baselines}} & \multicolumn{4}{c|}{\textbf{Few-shot AD Methods (Non-CLIP-based)}} & \multirow{3}*{\makecell{\textbf{ResAD++} \\ \textbf{(W50)}}} & \multicolumn{3}{c|}{\textbf{CLIP-based AD Methods}} & \multirow{3}*{\makecell{ResAD$^{\dag}$ \\ NeurIPS2024}} & \\
   &  & \makecell{RDAD \\ CVPR2022}& \makecell{UniAD \\ NeurIPS2022} & SPADE & PaDiM & \makecell{PatchCore \\ CVPR2022} & \makecell{RegAD \\ ECCV2022} &  & \makecell{WinCLIP \\ CVPR2023} & \makecell{AnomalyCLIP \\ ICLR2024} & \makecell{InCTRL \\ CVPR2024} & & \multirow{-2}*{\textbf{ResAD$^{\dag}$++}} \\
\midrule
 \multirow{9}*{\textbf{2-shot}} & MVTecAD & 61.2 & 61.4 & 79.6 & 81.4 & 77.2 & 82.5 & \underline{86.0} & 84.6 & 81.4 & - & 89.7 & \textbf{90.8} \\
 
    & VisA & 49.4 & 39.5 & 63.9 & 55.8 & 63.1 & 67.6 & 82.4 & 80.6 & \textbf{87.0} & - & 84.9 & \underline{86.6}  \\
  
   & BTAD & 70.7 & 43.5 & 70.4 & 70.8 & 60.7 & 75.9 & \textbf{77.4} & 66.9 & 74.8 & - & 73.8 & \underline{76.0}  \\
  
   & MVTec3D & 79.0 & 69.0 & 80.6 & 84.5 & 81.1  & 85.3 & \underline{90.0} & 88.8 & 86.4 & - & 89.9 & \textbf{90.7} \\

   & MPDD & 53.7 & 55.1 & 81.0 & 70.4 & 75.0  & 78.9 & 87.6 & 88.5 & \underline{88.7} & - & 92.2 & \textbf{92.8} \\

   & MVTecLOCO & 44.6 & 41.9 & \underline{61.0} & 53.0 & 51.7  & 57.5 & \textbf{61.2} & 57.5 & 54.3 & - & 60.2 & 60.8 \\

   & BraTS & 33.2 & 56.7 & \underline{73.1} & 65.1 & 72.3  & 41.1 & 68.2 & 70.8 & 68.5 & - & 72.0 & \textbf{74.6} \\

   & Kvasir-Seg & 24.8 & 20.8 & 32.8 & 25.3 & 29.0 & 18.6 & 32.2 & 36.4 & \underline{45.6} & - & 57.9 & \textbf{59.3} \\

    & \textbf{Average} & 52.1 & 48.5 & 67.8 & 63.4 & 63.8 & 63.4 & 73.1 & 71.8 & \underline{73.3} & - & 77.6 & \textbf{79.0} \\

 \midrule
 
  \multirow{9}*{\textbf{4-shot}} & MVTecAD & 61.2 & 61.4 & 84.8 & 85.2 & 81.2 & 86.7 & \underline{88.7} & 85.5 & 81.4 & - & 90.4 & \textbf{91.3} \\
 
    & VisA & 49.4 & 39.5 & 68.2 & 65.0 & 70.6 & 72.4 & 84.0 & 80.8 & \underline{87.0} & - & 86.9 & \textbf{87.8}  \\
  
   & BTAD & 70.7 & 43.5 & 73.3 & 74.4 & 67.2 & \underline{77.0} & \textbf{80.4} & 66.9 & 74.8 & - & 73.6 & 75.7  \\
  
   & MVTec3D & 79.0 & 69.0 & 84.2 & 85.8 & 84.1  & 88.4 & \underline{91.8} & 89.5 & 86.4 & - & 91.2 & \textbf{91.9} \\

   & MPDD & 53.7 & 55.1 & 86.5 & 76.9 & 85.4  & 80.4 & \underline{90.4} & 89.5 & 88.7 & - & 93.8 & \textbf{94.2} \\

   & MVTecLOCO & 44.6 & 41.9 & \textbf{62.9} & 56.9 & 55.4  & 58.2 & 61.9 & 58.5 & 54.3 & - & 61.0 & \underline{62.1} \\

   & BraTS & 33.2 & 56.7 & \underline{77.3} & 73.9 & 77.9  & 52.1 & 75.8 & 71.6 & 68.5 & - & 75.5 & \textbf{77.4} \\

   & Kvasir-Seg & 24.8 & 20.8 & 34.0 & 25.4 & 32.6  & 23.6 & 34.0 & 37.9 & \underline{45.6} & - & 59.1 & \textbf{60.0} \\

    & \textbf{Average} & 52.1 & 48.5 & 71.4 & 67.9 & 69.3 & 67.4 & \underline{75.9} & 72.5 & 73.3 & - & 78.9 & \textbf{80.1} \\
    
 \midrule

 \multirow{9}*{\textbf{8-shot}} & MVTecAD & 61.2 & 61.4 & 86.8 & 87.9 & 83.2 & 89.1 & \underline{90.1} & 86.1 & 81.4 & - & 91.4 & \textbf{92.2} \\
 
    & VisA & 49.4 & 39.5 & 72.3 & 69.7 & 68.4 & 75.2 & 86.7 & 79.9 & \underline{87.0} & - & 88.3 & \textbf{88.7}  \\
  
   & BTAD & 70.7 & 43.5 & 76.9 & 75.7 & 68.2 & \underline{77.8} & \textbf{80.2} & 66.7 & 74.8 & - & 74.0 & 75.9  \\
  
   & MVTec3D & 79.0 & 69.0 & 85.6 & 87.5 & 81.7  & 89.5 & \textbf{93.0} & 89.8 & 86.4 & - & 91.6 & \underline{92.3} \\

    & MPDD & 53.7 & 55.1 & 87.8 & 80.1 & 84.8  & 85.9 & \underline{92.5} & 89.7 & 88.7 & - & 94.5 & \textbf{94.9} \\

   & MVTecLOCO & 44.6 & 41.9 & \textbf{64.1} & 62.5 & 56.7  & \underline{63.5} & 63.0 & 58.8 & 54.3 & - & 61.7 & 63.1 \\
   
   & BraTS & 33.2 & 56.7 & \textbf{78.9} & 78.8 & \textbf{78.9}  & 58.6 & \underline{78.5} & 74.1 & 68.5 & - & 77.6 & \textbf{78.9} \\

   & Kvasir-Seg & 24.8 & 20.8 & 33.4 & 30.8 & 34.6 & 25.0 & 34.1 & 38.3 & \underline{45.6} & - & 60.6 & \textbf{61.0} \\

    & \textbf{Average} & 52.1 & 48.5 & 73.2 & 71.6 & 69.6 & 70.6 & \underline{77.3} & 72.9 & 73.3 & - & 80.0 & \textbf{80.9} \\
  
\bottomrule[0.5mm]
\end{tabular}}
\end{table*}

\begin{table*}[ht]
\caption{\textbf{Per-class results on the MVTecAD dataset}. $\cdot/\cdot$ means image-level AUROC and PRO.}
\label{tab:per_class_results}
\resizebox{1.0\linewidth}{!}{
\begin{tabular}{c|c||cc|cccc|c|ccc|a}
\toprule[0.5mm]
  \multirow{3}*{\textbf{Dataset}} & \multirow{3}*{\textbf{Classes}} & \multicolumn{2}{c|}{\textbf{Baselines}} & \multicolumn{4}{c|}{\textbf{Few-shot AD Methods (Non-CLIP-based)}} & \multirow{3}*{\makecell{\textbf{ResAD++} \\ \textbf{(W50)}}} & \multicolumn{3}{c|}{\textbf{CLIP-based AD Methods}} &  \\
   &  & \makecell{RDAD \\ CVPR2022}& \makecell{UniAD \\ NeurIPS2022} & SPADE & PaDiM & \makecell{PatchCore \\ CVPR2022} & \makecell{RegAD \\ ECCV2022} &  & \makecell{WinCLIP \\ CVPR2023} & \makecell{AnomalyCLIP \\ ICLR2024} & \makecell{InCTRL \\ CVPR2024} & \multirow{-2}*{\textbf{ResAD$^{\dag}$++}}\\
\midrule
 \multirow{15}*{\textbf{MVTecAD}} & Bottle & 70.9/35.3 & 61.6/51.3 & 94.0/93.7 & \underline{99.7}/\underline{94.8} & 99.6/91.5 & 99.3/91.9 & \textbf{99.9}/\textbf{95.5} & 99.5/85.3 & 89.3/80.5 & 99.2/- & 99.5/94.4 \\
 
    & Cable & 64.4/56.6 & 42.6/34.0 & 77.8/76.1 & 76.3/73.9 & \underline{93.8}/\underline{84.3} & 81.2/82.6 & 83.7/80.6 & 90.6/73.0 & 69.8/64.4 & 86.5/- & \textbf{97.3}/\textbf{85.6}  \\
  
   & Capsule & 46.6/76.6 & 45.1/67.2 & 68.6/\underline{88.7} & 63.8/87.9 & 67.6/82.5 & 65.9/87.9 & 66.9/\textbf{91.2} & 79.7/83.6 & \textbf{89.9}/87.2 & \underline{84.0}/- & 71.5/84.0  \\
  
   & Carpet & 99.5/90.2 & 98.0/96.3 & 89.4/95.3 & 99.8/\underline{97.5} & 98.3/92.4  & 98.2/94.7 & 98.9/95.4 & \underline{99.9}/95.5 & \textbf{100}/90.1 & 99.7/- & \textbf{100}/\textbf{97.6} \\

   & Grid & 70.7/22.0 & 90.1/69.2 & 30.6/55.0 & 68.1/64.4 & 62.1/37.9  & 85.8/68.4 & 88.1/70.9 & \underline{98.2}/\underline{84.6} & 97.0/75.6 & 97.8/- & \textbf{100}/\textbf{93.7} \\

   & Hazelnut & 85.5/88.4 & 82.6/75.0 & 75.6/91.7 & 96.6/91.3 & \underline{99.0}/87.2  & 93.5/92.2 & 97.6/\underline{93.3} & 98.0/\underline{93.3} & 97.2/92.4 & 95.4/- & \textbf{99.9}/\textbf{95.8} \\

   & Leather & 84.4/84.1 & \underline{99.8}/97.7 & 94.7/97.4 & \textbf{100/}98.0 & \underline{99.8}/92.8  & 97.2/96.3 & \textbf{100}/\underline{98.2} & \textbf{100}/98.0 & \underline{99.8}/92.2 & \textbf{100}/- & \textbf{100}/\textbf{98.3} \\

   & Metal\_nut & 64.7/65.8 & 48.2/28.9 & 56.1/84.1 & 60.5/66.4 & 90.2/85.8  & 88.0/88.3 & 94.3/\textbf{90.2} & \underline{96.8}/77.3 & 93.6/71.0 & 94.8/- & \textbf{100}/\underline{89.9} \\

    & Pill & 63.7/77.8 & 47.2/63.9 & 61.6/92.7 & 61.7/92.3 & 80.3/85.2 & 63.9/88.7 & 91.6/\textbf{96.6} & \underline{91.7}/90.6 & 81.8/88.2 & 89.0/- & \textbf{98.7}/\underline{96.3} \\

     & Screw & 62.6/79.4 & 51.9/71.5 & 44.3/81.1 & 49.7/77.5 & 51.3/67.7 & 58.1/87.1 & 58.0/86.1 & \underline{81.5}/86.3 & 81.1/\underline{88.0} & 80.0/- & \textbf{85.5}/\textbf{91.9} \\

      & Tile & 83.0/53.0 & 88.7/69.7 & 94.2/81.0 & 96.8/81.8 & 99.5/85.4 & 95.0/78.4 & \underline{99.8}/\underline{88.1} & 99.4/78.2 & \textbf{100}/87.6 & 99.9/- & 99.4/\textbf{89.6} \\

       & Toothbrush & 46.3/66.2 & 58.5/53.1 & 66.4/86.0 & 89.2/87.9 & 80.0/70.6 & 86.0/84.7 & \underline{99.7}/\textbf{93.1} & 95.0/\underline{89.4} & 84.7/88.5 & 97.5/- & \textbf{100}/\textbf{93.1} \\
       
       & Transistor & 52.9/31.9 & 51.8/19.9 & 89.2/64.7 & 81.0/\textbf{81.6} & \underline{95.4}/75.4 & 76.2/\underline{81.3} & 88.7/65.8 & 90.9/68.1 & 92.8/52.8 & 89.8/- & \textbf{96.5}/72.6 \\
       
       & Wood & 90.9/70.6 & 95.0/85.4 & 97.3/91.4 & 99.3/91.7 & 98.8/89.1 & 99.4/89.8 & \underline{99.7}/\textbf{93.6} & \textbf{99.8}/87.5 & 96.8/91.2 & 99.7/- & \textbf{99.6}/\underline{93.2} \\

       & Zipper & 69.6/20.6 & 61.7/37.6 & 92.8/\underline{92.9} & 87.8/90.3 & 95.0/90.0 & 83.8/87.9 & 95.1/92.5 & 98.0/91.4 & \underline{98.5}/65.3 & 96.5/- & \textbf{98.6}/\textbf{93.1} \\
\bottomrule[0.5mm]
\end{tabular}}
\end{table*}

\textbf{Setup.} Different from the conventional unsupervised anomaly detection, we train AD models on one dataset and then evaluate the model's class-generalization ability on another dataset. Specifically, we train ResAD++ on the MVTecAD dataset and evaluate it on other datasets. As for MVTecAD, we train ResAD++ on the VisA dataset. During training, the anomalies that exist in the training dataset will also be effectively utilized. As these anomalies belong to known classes, they will not cause any anomaly leakage of the test set into training. Due to page limitation, we report dataset-level results, which are averaged values across all respective classes in the dataset, with the number of few-shot normal samples set to $N_{fs}=2,4,8$.

\textbf{Competing Methods.} We select the representative single-class AD method (RDAD \cite{RDAD}) and the multi-class AD method (UniAD \cite{UniAD}) as baselines. Our method is mainly compared with few-shot AD methods. Following WinCLIP \cite{WinCLIP}, we adapt three conventional full-shot AD methods, including SPADE \cite{SPADE}, PaDiM \cite{PaDiM}, and PatchCore \cite{PatchCore}, to the few-shot setting by making use of few-shot normal samples to calculate distance-based anomaly scores. We also compare with the few-shot AD method RegAD \cite{RegAD}. These methods are all based on WideResNet50 to extract features. For a fair comparison, we also utilize WideResNet50 \cite{WideResNet} as the feature extractor. The corresponding model is denoted as ResAD++(W50). However, these methods still need to remodel in new classes based on few-shot normal samples (see discussions in Sec.\ref{sec:related_work}), while our ResAD++ can be directly applied to new classes only requiring extracting features of few-shot normal samples as reference. Then, we also compare with the recent CLIP-based AD methods, including WinCLIP \cite{WinCLIP}\footnote{No official implementation of WinCLIP is available. We use the public implementation at \url{https://github.com/zqhang/Accurate-WinCLIP-pytorch}.} and InCTRL \cite{InCTRL}, and a CLIP-based zero-shot AD method AnomalyCLIP \cite{AnomalyCLIP}. To guarantee the rationality of result comparison, we ensure all methods use the same few-shot normal samples, and all results are evaluated based on 224$\times$224 resolution. 

\subsection{Main Results}

Tab.\ref{tab:main_results} shows the comparison results of our ResAD++ and other SOTA competing methods in image-level AUROC and pixel-level AUROC, respectively, on eight real-world AD datasets. The PRO scores are shown in Tab.\ref{tab:main_results2}. Note that all the results are dataset-level average results across their respective data subsets. Compared to the results on trained classes (results in the original papers), our results (first two columns) demonstrate that conventional AD methods will fail when dealing with novel classes, whether it is the single-class (RDAD) or the multi-class (UniAD) AD method.


By comparison, we can see that our ResAD++(W50) can significantly outperform all non-CLIP-based AD methods on all the 2-shot, 4-shot, and 8-shot settings. With more few-shot normal images, the performance of all methods generally becomes better. On average, our ResAD++(W50) outperforms the best competing model, PatchCore, with up to 8.1\%/2.2\%/9.3\%, 5.1\%/0.9\%/6.6\%, and 4.7\%/1.2\%/7.7\% improvements under the 2-shot, 4-shot, and 8-shot settings, respectively. Please note that when evaluating PatchCore, we utilize the few-shot normal samples to remodel the coreset for each new class (see Sec.\ref{sec:related_work}), while our ResAD++(W50) is directly applied to each new class without any remodeling or fine-tuning. Even with remodeling, our method still has advantages over the conventional few-shot AD methods in cross-dataset generalization. Compared to the recent CLIP-based AD methods, our ResAD++(W50) by only using WideResNet50 can achieve comparable or even better results than WinCLIP and InCTRL (with more powerful ViT-B/16+) and also AnomalyCLIP (with ViT-L/14), further demonstrating our superiority.



\begin{table*}[ht]
\caption{\textbf{Framework ablation studies}. I-AUROC and P-AUROC mean image-level AUROC and pixel-level AUROC, respectively. ``RF'' means residual features, ``FHC'' means feature hypersphere constraining. ``AI-OCC Loss'' adopts the abnormal-invariant OCC loss. ``FDM'' represents feature distribution matching. ``MAC'' means the merged anomaly criterion (see descriptions in Sec.\ref{sec:anomaly_scoring}).}
\label{tab:ablation_framework}
\centering
\resizebox{1.0\linewidth}{!}{
\begin{tabular}{cccccc||ccc|ccc}
\toprule[0.5mm]
  \multirow{2}*{\textbf{expID}} & \multirow{2}*{\textbf{RF}} & \multirow{2}*{\textbf{FHC}} & \multirow{2}*{\textbf{AI-OCC Loss}} & \multirow{2}*{\textbf{FDM}} & \multirow{2}*{\textbf{MAC}} & \multicolumn{3}{c|}{\textbf{MVTecAD}} & \multicolumn{3}{c}{\textbf{VisA}}\\
& & & & & & I-AUROC & P-AUROC & PRO & I-AUROC & P-AUROC & PRO \\
\midrule
 3.1 & & &  &  & & 70.6 & 78.5 & 62.2 & 59.0 & 83.6 & 53.0 \\

 3.2 & \checkmark & &  &  & & 82.2 & 94.0 & 85.7 & 84.9 & 95.8 & 78.6\\

 3.3 & & &  & \checkmark & & 66.6 & 76.6 & 60.0 & 55.2 & 83.0 & 50.1\\

3.4 & \checkmark &  &  & \checkmark & & 87.2 & 94.6 & 86.2 & 85.3 & 96.0 & 81.4 \\

3.5 & \checkmark & \checkmark & \checkmark &  & & 89.3 & 95.3 & 87.1 & 85.5 & 96.8 & 82.1 \\

 3.6 & \checkmark & \checkmark &  &  & & 78.0 & 90.1 & 73.6 & 77.4 & 93.2 & 72.4 \\

 3.7 & \checkmark & \checkmark & \checkmark &  \checkmark & & 90.4 & 95.6 & 88.4 & 88.7 & 96.7 & 83.5\\

 3.8 & \checkmark & \checkmark & \checkmark &  \checkmark & \checkmark & 90.8 & 95.8 & 88.7 & 89.3 & 96.8 & 84.0 \\
 
\bottomrule[0.5mm]
\end{tabular}}
\end{table*}

 We further implement a ResAD$^{\dag}$++ model by utilizing the powerful ImageBind \cite{ImageBind} as the feature extractor. The outputs from the $[8, 16, 24, 32]$ layers of ImageBind are used as the pre-trained features. As shown in Tab.\ref{tab:main_results}, by employing a model with stronger representation capability, our method can achieve better cross-dataset performance, which significantly outperforms the SOTA CLIP-based AD methods, WinCLIP and InCTRL. This demonstrates that our framework can effectively combine the latest vision models to manifest a stronger class-adaptive ability. Moreover, these two CLIP-based methods also heavily rely on CLIP-based image encoders. When we employ WideResNet50 in these two methods, our method has more advantages than these two methods. The results under the 4-shot setting are in Tab.\ref{tab:few_shot_w50}. Compared to WinCLIP and InCTRL, our method is less reliant on the representation capability of the backbone network and is more widespreadly applicable for various backbones.

What’s more, when applied to medical AD datasets (generalization no longer at the class-level but at the domain-level), our method further demonstrates stronger cross-domain generalization ability, despite it being trained on industrial data (MVTecAD). Specifically, under the 4-shot setting, our method surpasses WinCLIP by 18.4\%/3.2\%/5.8\% (on BraTS) and 2.9\%/8.2\%/22.1\% (Kvasir-Seg) in image-level AUROC/pixel-level AUROC/PRO, respectively. Compared to the results of cross-class generalization in the industrial datasets, the cross-domain results also show that our method has more significant advantages in the more challenging task of cross-domain generalization. In Fig.\ref{fig:vis_results}, we show the qualitative cross-domain generalization results on the BraTS dataset in the last row. An obvious observation is that our method can locate anomalies more accurately and can also effectively avoid false positives in normal regions. By comparison, the previous AD methods either can’t correctly localize anomalies (the left example) or may still generate many normal misdetections (the right example).

 \textbf{Sensitivity discussion}. The extensive results on 8 real-world datasets have demonstrated that our method has strong cross-class and even cross-domain generalization capabilities. We also utilize VisA for training and MVTecAD for testing. With only 8-shot normal samples as reference, our method can achieve 98.6\% image-level AUROC and 96.7\% pixel-level AUROC on MVTecAD. This also indicates that our method does not rely on MVTecAD for training. For each dataset, we conduct experiments under the 2, 4, and 8 shot settings, and the results show that our method also performs well even with 2 reference samples. Thus, our method overall has good robustness to both the dataset and the number of reference samples.

\textbf{Per-class results}. In Tab.\ref{tab:per_class_results}, we further provide per-class results on the most representative dataset, MVTecAD. The per-class results can provide more substantial evidence that our method outperforms other methods, as our method can achieve the best or second-best results in
13 classes out of 15 classes. 

\begin{table*}[ht]
\caption{\textbf{Anomaly detection and localization results (under 4-shot) with WideResNet50 as the feature extractor}. $\cdot/\cdot/\cdot$ means image-level AUROC, pixel-level AUROC, and PRO, respectively.}
\label{tab:few_shot_w50}
\resizebox{0.55\linewidth}{!}{
\begin{tabular}{c||cc|c}
\toprule[0.5mm]
  \textbf{Dataset} & WinCLIP & InCTRL & ResAD++(W50) \\
\midrule
 \textbf{MVTecAD} & 86.6/91.6/81.7 & 86.9/-/- & 90.8/95.8/88.7 \\

  \textbf{VisA} & 80.7/92.5/76.2 & 82.3/-/- & 89.3/96.8/84.0 \\

  \textbf{BTAD} & 87.7/93.7/65.4 & 90.4/-/- & 94.1/97.3/80.4 \\

  \textbf{MVTec3D} & 63.1/91.7/83.6 & 63.2/-/- & 71.1/97.6/91.8 \\
\bottomrule[0.5mm]
\end{tabular}}
\end{table*}

\subsection{Comparison with Full-shot Trained Models}

Previous experiments have verified the class-generalization ability of our ResAD++ under three few-shot conditions. However, in practical applications, it's usually not hard for us to obtain a sufficient number of normal samples for a new class. Therefore, we further wonder to know that how our ResAD++ performs compared to the AD models trained on full-shot normal samples, and how our ResAD++ performs when it can access full-shot normal samples. Then, for a new dataset, we retrain UniAD and RDAD on full-shot normal samples (for a class, full-shot means all normal samples provided by this class's training set) from all classes in this dataset (only one model is trained, namely under the multi-class AD paradigm). The results of full-shot trained UniAD and RDAD are shown in Tab.\ref{tab:full_shot}, which can be regarded as the expected reference performance for each dataset. By comparison, our method can achieve comparable results with AD models retrained on new classes based on full-shot normal samples, even if only 8 normal samples are used as reference. Furthermore, applying our method to the full-shot is also quite easy, we only need to extract features from full-shot normal samples as reference. However, the full-shot will bring too many normal reference features, leading to excessive computation costs in generating residual features. To this end, we follow the training-free global retrieval approach in \cite{CPR} to match some spatially-aligned samples for each input sample, and then generate specific reference features for each input sample. Specifically, we matched 10 reference samples for each input sample. The results in Tab.\ref{tab:full_shot} show that, with accessing the full-shot normal samples, our method's performance on new classes can be further improved, surpassing the full-shot trained UniAD and RDAD. This further verifies the superiority of our method for class-agnostic anomaly detection and demonstrates its potential to achieve the same effect as trained classes on new classes.

\begin{table*}[ht]
\caption{\textbf{Performance comparison with full-shot trained AD models}. RDAD and UniAD are trained with the full-shot normal samples (all normal samples in the training set) when used for new classes.}
\label{tab:full_shot}
\resizebox{0.7\linewidth}{!}{
\begin{tabular}{c||cc|cc}
\toprule[0.5mm]
  \textbf{Dataset} & \textbf{RDAD} & \textbf{UniAD} & \makecell{\textbf{ResAD++(W50)} \\ (8-shot)} & \makecell{\textbf{ResAD++(W50)} \\ (full-shot)}\\
\midrule
 \textbf{MVTecAD} & 95.7/96.3/91.0 & 96.5/97.0/90.6 & 93.1/96.4/90.1 & 98.2/97.0/92.5\\

\textbf{VisA} & 89.5/97.5/85.5 & 92.8/98.3/86.8 & 90.6/97.2/86.7 & 93.3/98.3/88.9\\

\textbf{BTAD} & 93.9/97.5/74.8 & 94.2/97.2/76.8 & 94.0/97.3/80.2 & 94.4/97.5/81.3\\

\textbf{MVTec3D} & 77.1/98.3/93.1 & 77.5/96.6/88.5 & 74.5/98.0/93.0 & 86.0/98.0/93.5\\

\bottomrule[0.5mm]
\end{tabular}}
\end{table*}

\subsection{Ablation Studies}
\label{sec:ablation_studies}

In ablation studies, we conduct experiments under the ``VisA to MVTecAD'' and ``MVTecAD to VisA'' cases and use WideResNet50 \cite{WideResNet} as the feature extractor. All the experiments are under the 4-shot setting.

\textbf{Residual Features.} As shown in Tab.\ref{tab:ablation_framework}, without residual features, the cross-dataset performance drops dramatically from 82.2\%/85.7\% to 70.6\%/62.2\% on MVTecAD and 84.9\%/78.6\% to 59.0\%/53.0\% on VisA. This verifies our confirmation that residual features are of vital significance for class-agnostic anomaly detection. Analogously, any method that can realize feature decorrelation to reduce the variations of new class distribution relative to known class distributions is also promising to achieve class-agnostic anomaly detection. 

\textbf{Feature Hypersphere Constraining.} The ablation study on the effectiveness of the feature hypersphere constraining is also in Tab.\ref{tab:ablation_framework}. The effectiveness (expID 3.2 v.s. 3.5) indicates that by further reducing the variations in the feature distribution and making the distribution of new classes more consistent with the learned distribution, we can achieve better cross-class generalization results and ultimately arrive at the general anomaly detection goal. In Fig.\ref{fig:vis_scale_decorrelation}, we also present a visualization figure to intuitively show the effect of the Feature Constraintor.


\textbf{Anomaly-invariant OCC Loss.} The effectiveness of anomaly-invariant OCC loss is validated in Tab.\ref{tab:ablation_framework}. With the anomaly-invariant OCC loss, image-level AUROC and PRO can be improved by 11.3\%/13.5\% on MVTecAD and 8.1\%/9.7\% on VisA (expID 3.5 v.s. 3.6). Moreover, we also find that without this loss, the results would rapidly decrease after certain epochs of training (\emph{i.e.}, overfitting). This shows that keeping abnormal residual features as invariant as possible is beneficial to avoid the Feature Constriantor overfitting and thus achieve better results. 

\begin{table*}[ht]
\caption{\textbf{Ablation studies of hyperparameters in the VQ-based FDM module}. To avoid excessive hyperparameter combinations, we fix the $K$ as 1024 when we change the $\alpha$. The $\alpha$ is set as 0.4 when the $K$ is changing.}
\label{tab:ablation_fdm}
\resizebox{1.0\linewidth}{!}{
\begin{tabular}{c|ccccc|ccccc}
\toprule[0.5mm]
& \multicolumn{5}{c|}{Feature distribution matching coefficient ($\alpha$, in Eq.(\ref{eq:fdm}))} & \multicolumn{5}{c}{Number of codebook embeddings ($K$)} \\
     & 0.3 & 0.4 & 0.5 & 0.6 & 0.7 & 512 & 1024 & 1536 & 2048 & 2560\\
\midrule
   \textbf{MVTecAD} & 89.8/95.2/87.9 & 90.7/95.7/88.7 & 90.6/95.7/88.6 & 90.6/95.6/88.5 & 90.4/95.5/88.4 & 90.6/95.7/88.5 & 90.7/95.7/88.7 & 90.8/95.8/88.7 & 90.5/95.6/88.5 & 90.6/95.6/88.4\\
 
    \textbf{VisA} & 89.3/96.4/83.8 & 89.3/96.8/83.9 & 89.1/96.7/83.9 & 88.5/96.8/83.8 & 87.9/96.7/83.2 & 89.0/96.7/83.7 & 89.3/96.8/83.9 & 89.3/96.8/84.0 & 88.5/96.8/83.9 & 88.5/96.6/83.8 \\
    
\bottomrule[0.5mm]
\end{tabular}}
\end{table*}

\textbf{Feature Distribution Matching.} As shown in Tab.\ref{tab:ablation_framework}, with feature distribution matching, the cross-dataset performance can be further improved by 1.1\%/1.3\% on MVtecAD and 2.2\%/1.4\% on VisA (expID 3.5 v.s. 3.7). The results demonstrate that FDM is effective in improving the class-generalization capability of our ResAD. However, directly applying FDM to the initial features (expID 3.3), the results are not improved but decreased. This may be because when the discrepancy between testing and training distributions is too large, it's hard for FDM to reduce the distribution mismatch under such a condition. In initial features, the discrepancy between testing and training distributions will be significantly greater than the discrepancy in residual features. Residual features are more conducive to FDM's effectiveness. We think that FDM will be a promising technique for class-agnostic anomaly detection, deserving more explorations in future work.

\textbf{Feature Constraintor Configuration.} We further ablate the network architectures of the Feature Constraintor, the results are shown in Tab.\ref{tab:ablation_constraintor}. The results indicate that the simple Conv+BN+ReLU network can yield the best performance. We observe a significant performance drop with a more complex feature constraintor (\emph{e.g.}, Bottleneck, MultiScaleFusion). One possible reason is that a complex network may lead to overfitting, reducing the generalization ability for various anomalies in the test.

\textbf{Hyperparameters of the VQ-based FDM module.} We ablate the hyperparameters $\alpha$ and $K$ in Tab.\ref{tab:ablation_fdm}. From Tab.\ref{tab:ablation_fdm}, we can draw the following main conclusions: (1) $\alpha$ has a more significant effect on performance compared to $K$. Small $\alpha$ may lead to excessive training distribution injecting, while large $\alpha$ may result in ineffective matching with the training distribution, both of which can cause performance degradation. Overall, setting $\alpha$ to 0.4 is a good choice. (2) The VQ-based FDM module is not very sensitive to $K$, setting $K$ to a moderate number (1536) can achieve the best results. 


\begin{table*}[ht]
\caption{\textbf{Comparison of different feature constraintors}. ``ConvBnRelu'' implements a simple Conv+BN+ReLU network. ``BasicBlock'' adopts the BasicBlock in ResNet at each feature level. ``BottleNeck`` replaces the BasicBlock with BottleNeck. ``MultiScaleFusion'' is a FPN-like architecture to fuse multi-scale features. In ``MultiScaleFusion+BasicBlock/BottleNeck'', we add BasicBlock/BottleNeck after the multi-scale fusion.}
\label{tab:ablation_constraintor}
\resizebox{0.55\linewidth}{!}{
\begin{tabular}{c||c|c}
\toprule[0.5mm]
  \textbf{Network Architecture} & \textbf{MVtecAD} & \textbf{VisA} \\
\midrule
ConvBnRelu & 90.8/95.8/88.7 & 89.3/96.8/84.0 \\ 

 BasicBlock & 87.8/94.9/87.4 & 76.4/92.9/72.8 \\

BottleNeck & 85.7/94.6/86.8 & 75.8/92.0/70.6 \\

MultiScaleFusion & 86.6/94.5/86.1 & 75.6/93.2/69.0 \\

 MultiScaleFusion+BasicBlock & 85.2/94.4/86.1 & 74.2/91.6/68.3 \\

  MultiScaleFusion+BottleNeck & 85.7/94.2/85.7 & 73.8/90.6/66.8 \\
\bottomrule[0.5mm]
\end{tabular}}
\end{table*}

\subsection{Generalization to Other AD Frameworks}
\label{sec:generalization_other_AD}
Furthermore, we think that our residual feature learning insight is not limited to the model proposed in this paper, but can be considered as an effective and general method for solving class-agnostic anomaly detection. The main reasons are: (1) The process of converting initial features to residual features can be easily applied to other AD models. (2) Residual features are less sensitive to new classes (see explanations in Sec.\ref{sec:residual_features}). In this subsection, we further extend our method to two popular AD frameworks: reconstruction-based and discrimination-based AD frameworks. Specifically, we employ UniAD \cite{UniAD} (reconstruction-based) and SimpleNet \cite{SimpleNet} (discrimination-based) as baselines and incorporate our method into them, the experimental results are shown in Tab.\ref{tab:other_framework}. It can be found that the performance of UniAD and SimpleNet is quite poor when used for new classes, while converting to residual feature learning can significantly improve the models' class-generalization capability. The remarkable improvements (\emph{e.g.}, 24.7\%/26.4\% and 15.9\%/23.5\% on MVTecAD) validate the effectiveness and universality of residual features for designing class-agnostic AD models.

\begin{table*}[ht]
\caption{\textbf{Anomaly detection and localization results when incorporating our method into UniAD and SimpleNet}. ``RFL'' represents residual feature learning.}
\label{tab:other_framework}
\resizebox{1.0\linewidth}{!}{
\begin{tabular}{c|cccc|c|cccc}
\toprule[0.5mm]
    &  MVTecAD & VisA & BTAD & MVTec3D & & MVtecAD & VisA & BTAD & MVTec3D \\
\midrule
   \textbf{UniAD} \cite{UniAD} & 68.2/80.5/61.4 & 49.9/79.7/39.5 & 69.7/82.4/43.5 & 52.2/89.1/69.0 &  \textbf{SimpleNet} \cite{RDAD} & 70.4/77.3/61.2 & 61.3/80.9/49.4 & 83.0/92.9/70.7 & 55.8/93.5/79.0\\
 
    + RFL (Ours) & 92.9/94.9/87.8 & 85.8/96.8/83.9 & 84.3/93.9/67.3 & 77.0/96.9/89.5 & + RFL (Ours) & 86.3/94.5/84.7 & 82.3/95.1/78.9 & 93.5/95.2/81.0 & 66.2/94.7/89.6  \\

   $\Delta$ & \textcolor{red}{+24.7/14.4/26.4} & \textcolor{red}{+35.9/17.1/44.4} & \textcolor{red}{+14.6/11.5/23.8} & \textcolor{red}{+24.8/7.8/20.5} &  $\Delta$ & \textcolor{red}{+15.9/17.2/23.5} & \textcolor{red}{+21/14.2/29.5} & \textcolor{red}{+10.5/2.3/10.3} & \textcolor{red}{+10.4/1.2/10.6}  \\
   
\bottomrule[0.5mm]
\end{tabular}}
\end{table*}

\subsection{Visualization and Qualitative Results}
\begin{figure}[ht]
    \centering
    \includegraphics[width=1.0\linewidth]{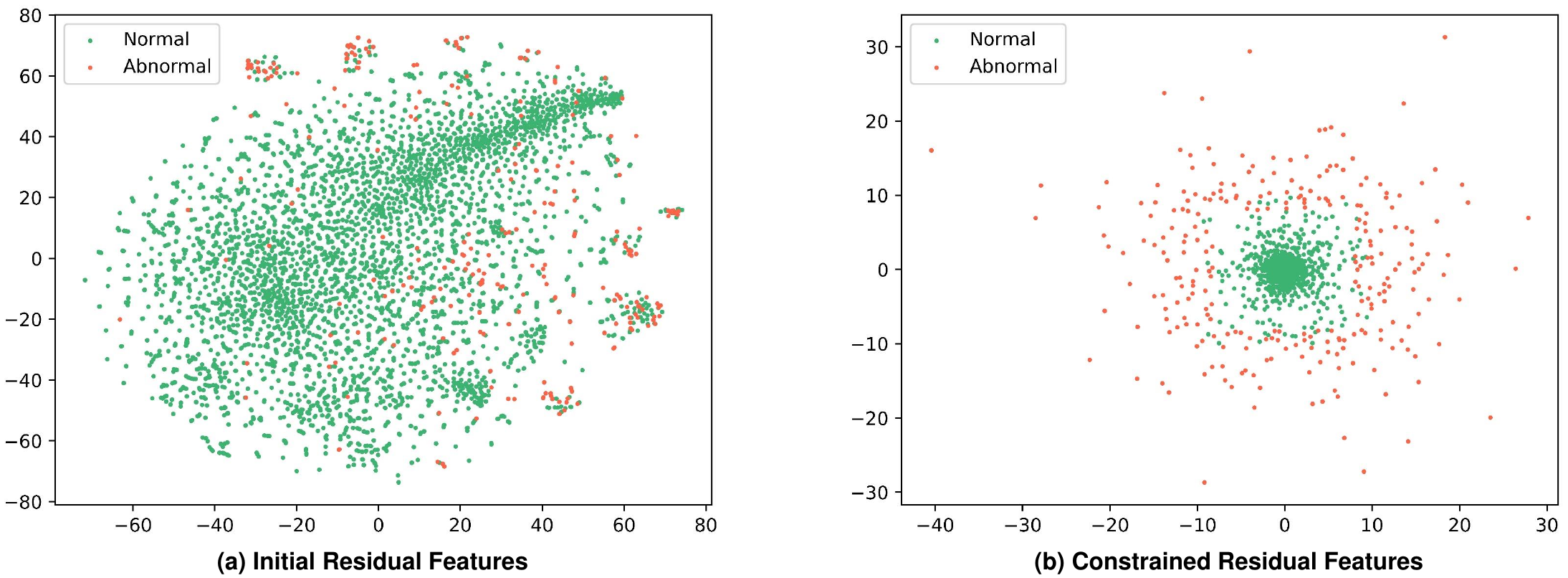}
    \caption{\textbf{Feature t-SNE visualization}. (a) The initial residual features. (b) The residual features after the Feature Constraintor.}
    \label{fig:vis_scale_decorrelation}
\end{figure}

\textbf{Visualization Results.} In Fig.\ref{fig:vis_feature_decorrelation}, we have shown the t-SNE visualization of initial features and residual features. It can be found that in the initial feature space, the feature distribution of new classes is significantly different from the distribution of known classes, resulting in poor adaptability of AD models to new classes. However, the variations among different classes can be significantly reduced by converting into residual feature space. In this way, the model's generalizability to new classes can be effectively improved. Fig.\ref{fig:vis_scale_decorrelation} (a) and (b) further show the t-SNE visualization of initial residual features and residual features after feature hypersphere constraining. Results show that the proposed feature hypersphere constraining approach can make the normal residual features more compact and more separated from the abnormal features.

\begin{figure*}[ht]
    \centering
    \includegraphics[width=1.0\linewidth]{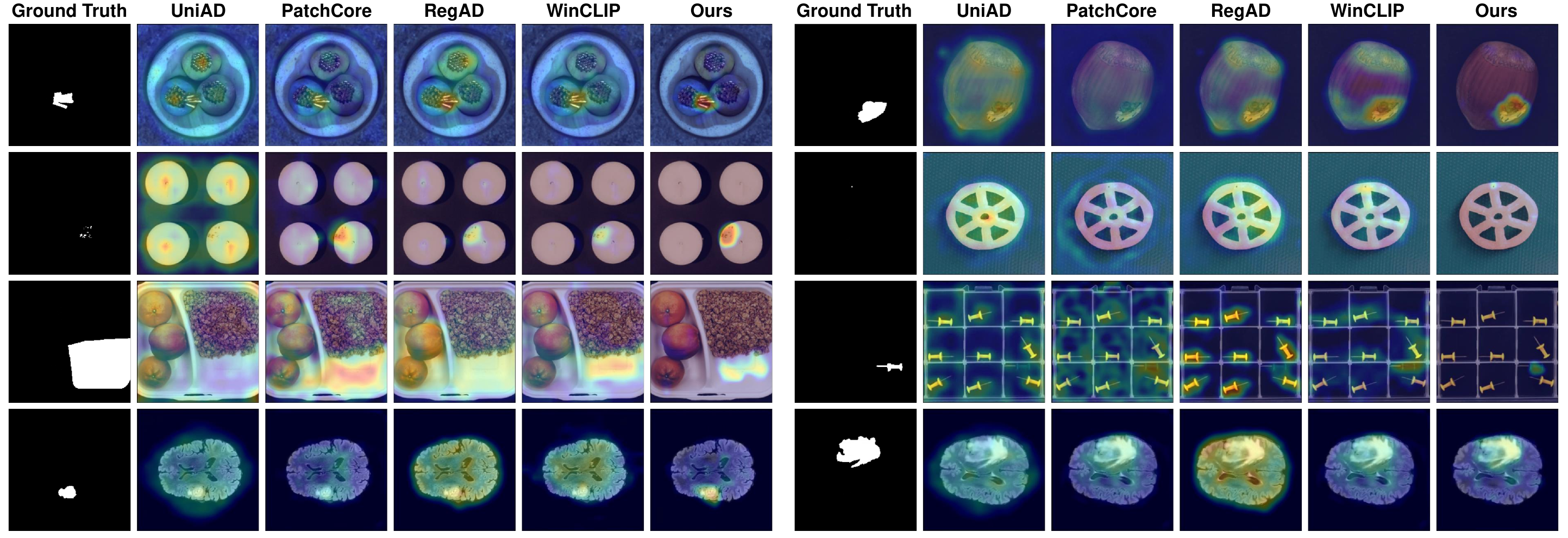}
    \caption{\textbf{Qualitative results across various classes from different datasets}. From the first to the last row, the samples are from the MVTecAD, VisA, MVTecLOCO, and BraTS datasets, respectively.}
    \label{fig:vis_results}
\end{figure*}

\textbf{Qualitative Results.} Fig.\ref{fig:vis_results} shows qualitative results across various classes from different datasets. From the first to the last row, the samples are from the MVTecAD, VisA, MVTecLOCO, and BraTS datasets, respectively. An obvious observation is that our method can locate anomalies more accurately and can also effectively avoid false positives in normal regions. However, the previous AD methods either can't correctly localize anomalies (\emph{e.g.}, the second samples in the second and third rows), or may still generate many normal misdetections (\emph{e.g.}, almost all visualized samples). 


\subsection{Computational Complexity}

\begin{figure}[ht]
    \centering
    \includegraphics[width=1.0\linewidth]{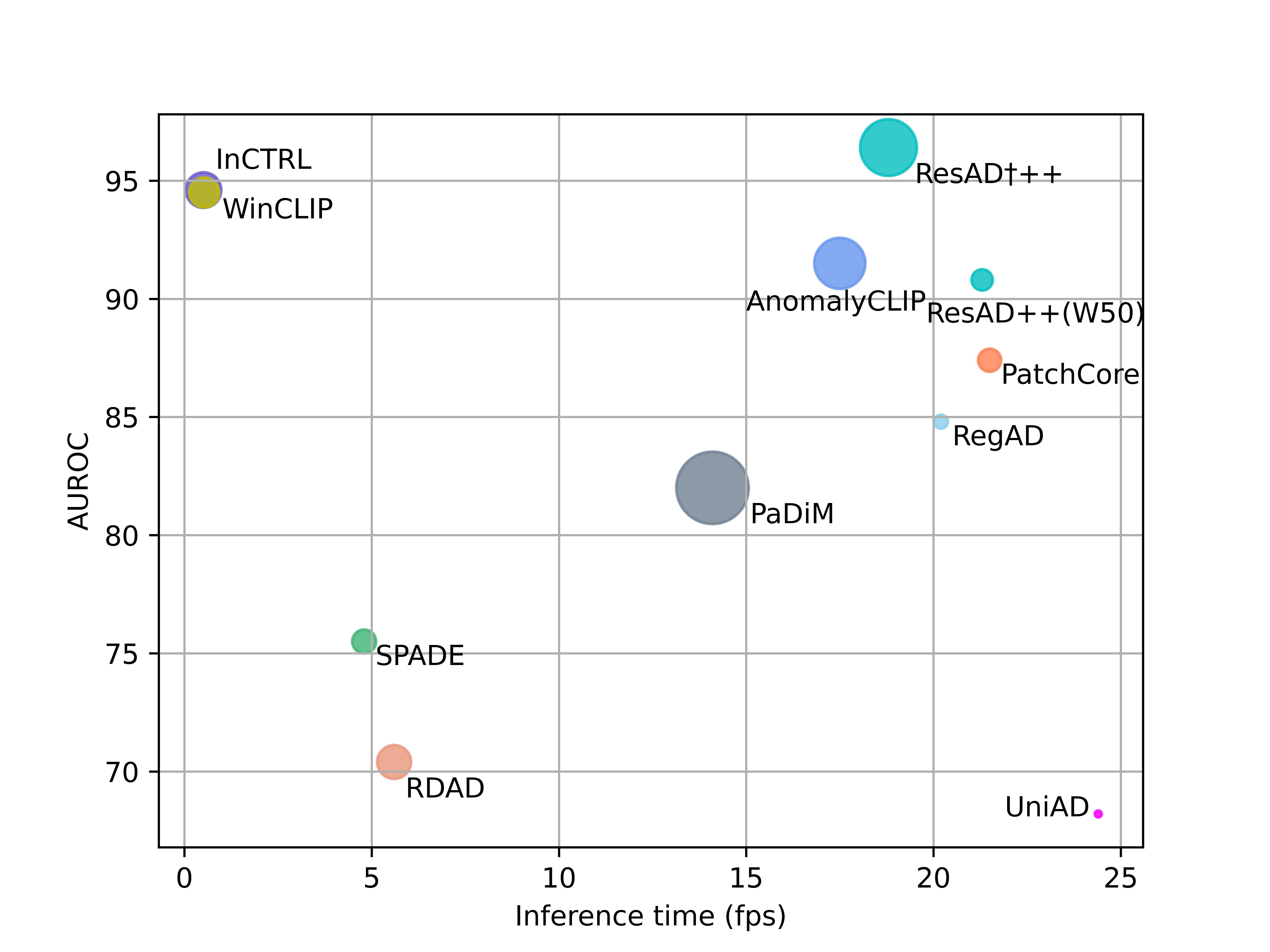}
    \caption{\textbf{The latency, AUROC performance versus model parameters on the MVTecAD dataset}. The AUROC is the image-level AUROC under the 4-shot setting. Our method can achieve a better latency, parameter/accuracy trade-off. The inference times are measured on a Nvidia RTX 4090 GPU with 24GB of VRAM.}
    \label{fig:computation_complexity}
\end{figure}

 With the image size fixed as $224\times224$, we compare the number of parameters and per-image inference time with all competitors. The comparison results are shown in Fig.\ref{fig:computation_complexity}. Compared to the conventional few-shot AD methods (SPADE, PaDiM, and PatchCore), our ResAD++(W50) overall has fewer parameters with the same backbone network (WideResNet50) as the feature extractor. Compared to WinCLIP and InCTRL, our ResAD++(W50) has fewer parameters and is significantly faster. Our ResAD$^{\dag}$++ also has the same magnitude of parameters, but its inference time is significantly faster than WinCLIP and InCTRL. Overall, our method can achieve a better latency, parameter/accuracy trade-off than these previous competing AD methods. 
 

\section{Discussion and Conclusion}
\textbf{Limitation.} Even if our method manifests good cross-class and even cross-domain generalization AD performance on eight real-world AD datasets, including industrial anomalies and medical anomalies, there are still some limitations of our work. One limitation is that we only conducted experiments on data of image modality, it's very valuable to further extend our method to other application domains and data modalities, such as video data and time series, to more comprehensively validate our method's generalizability. Another valuable future work is to incorporate our method into recent SOTA AD methods for achieving better class-agnostic AD performance. In Sec.\ref{sec:generalization_other_AD}, we incorporate our method into UniAD and SimpleNet and gain remarkable improvements. However, these methods still have a large room for improvement. How to upgrade the other types of anomaly detection methods to class-agnostic AD methods and how to find a general approach for class-agnostic (or even domain-agnostic) anomaly detection will be the future works.

In a summary, we innovatively propose the residual feature learning approach and further construct a simple but effective framework: ResAD++, for achieving class-agnostic anomaly detection. ResAD++ consists of several simple neural network modules that are easy to train and apply in real-world scenarios. Despite the simplicity, ResAD++ achieves remarkable AD results in new classes. We conclude our findings for future research: residual features are really effective for designing class-agnostic AD models, and our feature hypersphere constraining approach and the feature distribution matching techniques both have good reference values for future work.

\textbf{Data Availability.} The related code and data for ResAD++ framework are available at \url{https://github.com/xcyao00/ResAD}.

\section*{Acknowledgments}
This work was supported in part by the National Natural Science Fund of China (62371295), National Natural Science Fund of China (62201062), and the Science and Technology Commission of Shanghai
Municipality (22DZ2229005).

\bibliographystyle{IEEEtran}
\bibliography{sn-bibliography}

\end{document}